\documentclass[twocolumn]{article}

\usepackage[utf8]{inputenc} 
\usepackage[T1]{fontenc}    
\usepackage{hyperref}       
\usepackage{url}            
\usepackage{booktabs}       
\usepackage{amsfonts}       
\usepackage{microtype}      
\usepackage{textcomp}
\usepackage{float}

\usepackage{graphicx}
\usepackage{pdfpages}
\usepackage{caption}
\usepackage{subcaption}
\usepackage[a4paper,margin=1in,footskip=0.25in]{geometry}
\usepackage{amsmath, amssymb}

\usepackage[sort, numbers]{natbib}

\usepackage{xcolor}

\usepackage{nkj}
\usepackage{nkj_e}

\newcommand{\method}{L-Cool}

\newif\ifSuppressMemo
\ifSuppressMemo
\newcommand{\memo}[1]{}
\else
\usepackage{color}
\newcommand{\memo}[1]{{\bf \textcolor{red}{[#1]}}}

\newcommand{\reply}[1]{{\bf \textcolor{cyan}{[#1]}}}
\fi

\newcommand{\comm}[1]{}

\newcommand{\Obj}{L}

\title{Langevin Cooling for Domain Translation}

\author{
  Vignesh Srinivasan, 
  Klaus-Robert M{\"u}ller$^{\thanks{Corresponding authors: K.-R. M{\"u}ller, W. Samek and S. Nakajima. \newline
  V. Srinivasan is with the Machine Learning Group, Fraunhofer Heinrich Hertz Institute,
10587 Berlin, Germany. \newline (e-mail:
vignesh.srinivasan@hhi.fraunhofer.de). \newline 
W. Samek is with the Machine Learning Group, Fraunhofer Heinrich Hertz Institute,
10587 Berlin, Germany and also with BiFOLD. 
(e-mail: wojciech.samek@hhi.fraunhofer.de). 
\newline
K.-R. M{\"u}ller is with the Machine Learning Group, Technische Universit{\"a}t Berlin, 10587 Berlin, Germany, and also with BiFOLD and the Dept. of Brain and Cognitive Engineering, Korea University, Seoul 136-713, South Korea and Max Planck Institute for Informatics, 66123 Saarbr{\"u}cken, Germany. 
(e-mail: klaus-robert.mueller@tu-berlin.de). 
\newline
S. Nakajima is with the Machine Learning Group, Technische Universit{\"a}t Berlin, 10587 Berlin, Germany and also with BiFOLD and RIKEN AIP, 1-4-1 Nihonbashi, Chuo-ku, Tokyo
103-0027, Japan. 
(e-mail: nakajima@tu-berlin.de)
}}, \emph{Member, IEEE}$, \\
 Wojciech Samek$^{*}, \emph{Member, IEEE}$
 and Shinichi Nakajima$^{*}
$
}
\date{}

\begin{document}

\maketitle

\begin{abstract}
Domain translation is 
the task of finding correspondence between two domains.
Several Deep Neural Network (DNN) models, e.g., CycleGAN
and cross-lingual language models,
have 
shown remarkable successes on this task under the unsupervised setting---the mappings between the domains are learned from two independent sets of training data in both domains (without paired samples).
However, those methods typically do not perform well on a significant proportion of test samples.
In this paper,
we hypothesize that many of such unsuccessful samples lie at the \emph{fringe}---relatively low-density areas---of data distribution,
where the DNN was not trained very well,
and propose to perform Langevin dynamics to 
bring such fringe samples towards high density areas.
We demonstrate qualitatively and quantitatively that our strategy, 
called \emph{Langevin Cooling} (\method{}),
enhances state-of-the-art methods
in image translation
and language translation tasks.


\end{abstract}

\section{Introduction}
\label{sec:Inroduction}

Recently, Deep Neural Networks (DNNs) have broadly contributed across various  application domains in the sciences \cite{biswas2019prestack, gilmer2017neural, schutt2017schnet, schutt2019unifying, zhang2018deep, schmidt2019recent, arridge2019solving, bubba2019learning} 
and the industry \cite{codevilla2018end, dosovitskiy2017carla, AmazonCars, VoyageDeepDrive, GoogleFederatedLearning, wu2016google, MicrosoftGaming}. 
One of the notable successes is in 
unsupervised domain translation (DT),
on which this paper focuses.
DT is the task of translating data from a 
source domain to a target domain,
which has applications in super-resolution \cite{johnson2016perceptual}, language translation \cite{he2016dual, lample2019cross, edunov2018understanding}, image translation \cite{gatys2015neural, dong2015image, isola2017image, ulyanov2018deep}, text-image translation \cite{reed2016generative, zhang2017stackgan}, and data augmentation \cite{sandfort2019data, wu2018conditional, frid2018synthetic, bowles2018gan} among others.

\begin{figure}[!t]
\centering
\includegraphics[width=0.85\columnwidth, keepaspectratio]{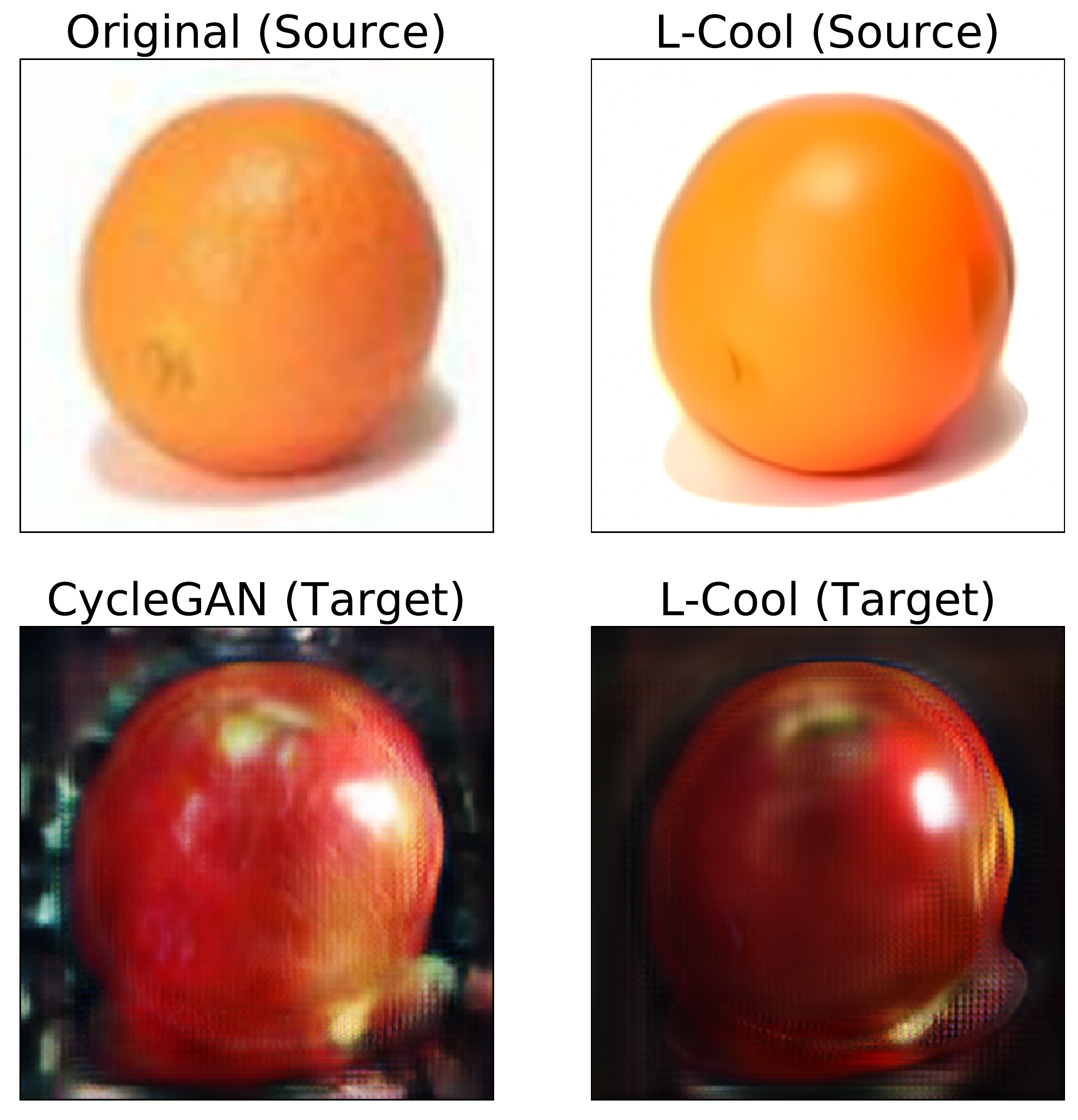}
\caption{
An example of orange2apple task.
The baseline
CycleGAN transfers an orange image to an apple image (left column).
Our proposed \method{} makes a slight change in the original orange image, which significantly improves the quality of the transferred apple image (right column):
the green artifacts surrounding the apple were removed almost completely,
and the texture and the color on the apple were improved,
although slight blurry along the edges of the apple was introduced.
}
\label{main_fig}
\vspace{-1mm}
\end{figure}

\begin{table}[t]
\caption{
Examples of French-English translation by XLM \cite{lample2019cross} 
and \method{}.
\method{} makes the translation closer to the ground-truth.
}
\resizebox{\columnwidth}{!}{
\begin{tabular}{c c }
    \toprule
     Original sentence & Le prix du pétrole continue à baisser \\ 
     & et se rapproche de 96 \$ le baril \\[1ex]
   {\bf  Ground-truth translation} & {\bf Oil extends drop toward \$ 96 a barrel} \\[1ex]
    XLM \cite{lample2019cross} (baseline) & Oil price continues to drop \\
    & and moves past \$ 96 a barrel  \\[1ex]
   {\bf \method{} (Ours)} & {\bf Oil price continues to drop} \\
     &{\bf and moves closer to \$ 96 a barrel} \\
    \midrule
    Original sentence & " Au milieu de XXe siècle , 
     on appelait cela une \\ 
     & urgence psychiatrique " , a indiqué Drescher\\[1ex]
     {\bf  Ground-truth translation} & {\bf " Back in the middle of the 20th century ,  it was} \\ & {\bf called a ' psychiatric emergency ' " said Drescher.} \\[1ex]
    XLM \cite{lample2019cross} (baseline) & " In the late 20th century , we called \\ & this a psychiatric emergency , " Drescher said\\[1ex]
  {\bf  \method{} (Ours)} & {\bf " In the middle of the 20th century , we called} \\ 
     & {\bf this a psychiatric emergency , " Drescher said .} \\
    \bottomrule
    \end{tabular}
    }
    \label{table:translation_enfr}
\vspace{-3mm}
\end{table}

In some DT applications,
labeled samples, i.e., paired samples in the two domains,
can be collected cheaply.  
For example,
in the super-resolution, a paired low resolution image can be created 
by artificially blurring and down-sampling a high resolution image.
However, in many other applications
including image translation and language translation,
collecting paired samples require significant human effort,
and thus only a limited amount of paired data are available.


\comm{
\begin{figure}[t!]
\centering
\includegraphics[width=\columnwidth]{files/journal_images/temp_xlm/val_enfr.pdf}
\caption{
\memo{Maybe better to send this to Experiment section?}
\reply{
Similar plot 
exists for the 
temp analysis. 
Do you think this would still
make sense to move it to Exp section?
}
The performance (BLEU score) in the EN-FR translation task on 
the validation set of the NewsCrawl dataset.
The proposed \method{} significantly improves the baseline performance by XML \cite{lample2019cross}.
\method{} was performed with DAE with training noise $\sigma^2 = 1$,
the temperature $T = 0.001$, and the number $N = 25$ of steps.
}
\label{main_fig}
\end{figure}
}

Unsupervised DT methods eliminate the necessity of paired data for supervision,
and only require independent sets of training samples in both domains.
In computer vision,
CycleGAN,
an extension of
Generative Adversarial Networks (GAN) \cite{goodfellow2014generative},
showed its capability of unsupervised DT with
impressive results in image translation tasks
\cite{zhu2017unpaired,kim2017learning,yi2017dualgan}.
It learns the mappings between the two domains by matching 
the source training distribution transferred to the target domain
and the target training distribution, under the cycle consistency constraint.  
Similar ideas were applied to natural language processing (NLP):
Dual Learning \cite{he2016dual,vaswani2017attention}
and
cross-lingual language models (XLM) \cite{lample2019cross},
which are trained on unpaired monolingual data,
achieved high performance in language translation.


 


Despite their remarkable successes, 
existing
unsupervised DT methods 
are known to fail on a significant proportion of test samples 
\cite{zhu2017unpaired, cycleganfailurecases, mejjati2018unsupervised}.
In this paper, 
we hypothesize that some of the unsuccessful samples are at the \emph{fringe} of the data distribution, i.e., they lie slightly off the data manifold, and therefore the DNN was not trained very well for translating those samples.
This hypothesis leads to our proposal
to bring fringe samples towards the high density data manifold,
where the DNN is well-trained,
by \emph{cooling down} the test distribution.
Specifically,
our proposed method, called \method{}, 
performs the Metropolis Adjusted Langevin Algorithm (MALA) to lower the temperature of test samples before applying the base DT method.
The gradient of the log-probability, which MALA requires, is estimated by the denoising autoencoder (DAE) \cite{Alain14}.




\method{} is generic and 
can be used for enhancing any DT method.
We demonstrate its effectiveness in image translation and language translation tasks,
where 
\method{} exhibits consistent performance gain.
Figure~\ref{main_fig} and
Table~\ref{table:translation_enfr}
show a few intuitive exemplar results.

\begin{figure*}[t]
\centering
\includegraphics[width=\textwidth]{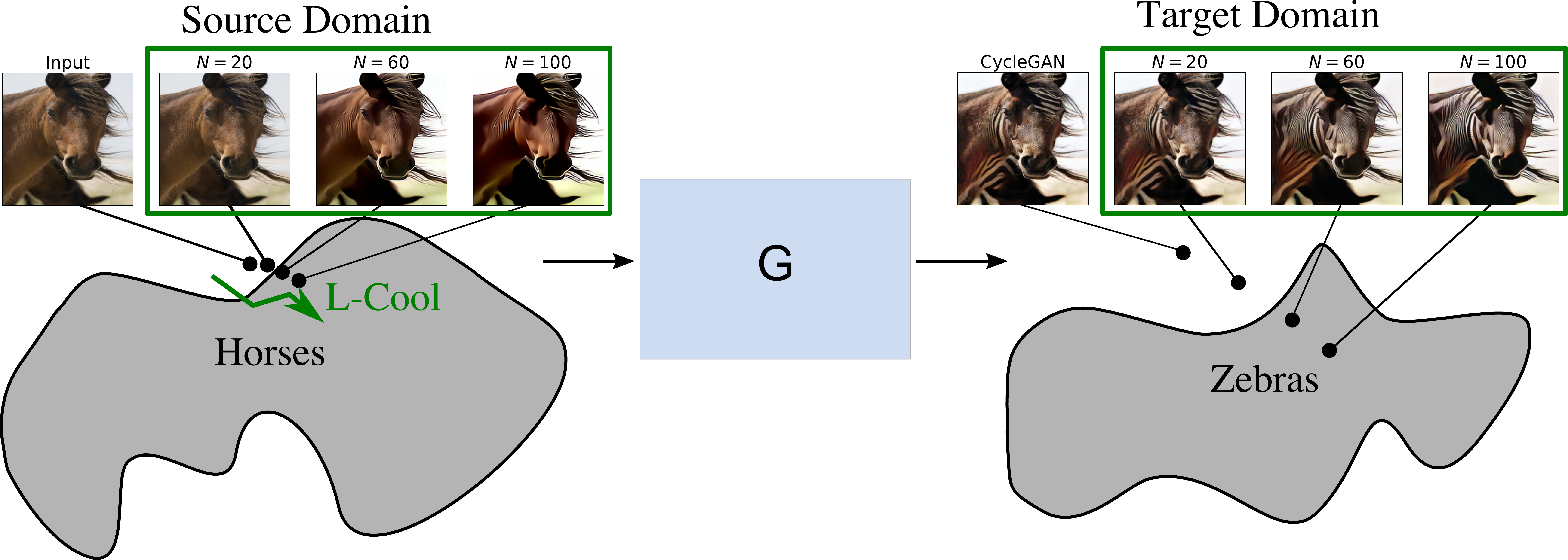} 
\caption{
\method{} drives the test sample in the source (horse) domain slightly towards the center of data manifold, which gives a significant impact on the translated sample in the target (zebra) domain.
}
\label{fig:cartoon}
\end{figure*}

This paper is an extension of our preliminary conference publication \cite{SriAAAI20} with the following new contributions:
\begin{itemize}
    \item 
    Evaluation in Language translation (English $\leftrightarrow$ French and English $\leftrightarrow$ German) on the NewsCrawl dataset\footnote{http://www.statmt.org/wmt14/index.html},
    which revealed  quantitative performance gain by \method{} in terms of the BLEU score \cite{papineni2002bleu}.
    
    \item   
    Comparison between the gradient estimators by DAE and by the cycle structure of CycleGAN.  The latter was mainly used in the conference version.
    
    \item 
    Analysis of hyperparameter dependence.
    
\end{itemize}

\subsection{Related Work} 
{
\subsubsection{{ Unsupervised Image Translation}}
CycleGAN \cite{zhu2017unpaired} and its concurrent works \cite{,kim2017learning,yi2017dualgan} 
have eliminated the necessity of supervision for image translation \cite{isola2017image,wang2018high}    
by using the loss inspired by GAN \cite{goodfellow2014generative} along with
the cycle-consistency loss.
The consistency requirement
forces translation to retain the contents of source images
so that they can be translated back.
\cite{liu2017unsupervised} proposed
a variant that shares the latent space between the two domains, which 
works as additional regularization for
alleviating the highly ill-posed nature of unsupervised domain translation.


\cite{huang2018multimodal} and \cite{lee2018diverse} 
tackled the general issue of unimodality in sample generation
by splitting
the latent space into two---a content space and a style space. 
The content space is shared between the two domains but the style space is unique to each domain. The style space is modeled with a Gaussian prior, which helps in generating diverse images at test time.
\cite{mejjati2018unsupervised, ugatit} 
showed that attention maps  
can boost the performance
by making the model focus on 
relevant regions in the image. 
Despite a lot of new ideas proposed for improving the image translation performance, 
CycleGAN \cite{zhu2017unpaired} is still considered to be the state-of-the-art in many transformation tasks.
}

\subsubsection{{ Unsupervised Language Translation}}
Language translation has been tackled with DNNs
with encoder-decoder architectures, 
where text in the source language is fed to the encoder 
and the decoder generates its translation in the target language
\cite{bahdanau2014neural}. 
Unsupervised language translation methods
have enabled 
learning from a large pool 
of monolingual data \cite{he2016dual, artetxe2017unsupervised},
which can be cheaply collected through the internet without any human labeling effort.

Transformers \cite{vaswani2017attention}
with attention mechanisms
have shown their excellent performance 
in unsupervised language translation,
as well as many other NLP tasks including language modelling,
understanding, and sentence classification. 
It was shown that 
generative pretraining strategies like 
Masked Language Modeling 
(which masks a portion of the words in the 
input sentence and forces the model to predict the masked words)
is effective in making transformers
better at language understanding
\cite{radford2018improving, radford2019language, devlin2018bert, lewis2019bart}.
Back translation 
has also enhanced
  performance 
by being a source of data augmentation 
while maintaining the 
cycle consistency constraint 
\cite{lample2017unsupervised, edunov2018understanding, poncelas2018investigating}. 
Cross-lingual language models
(XLM) 
\cite{lample2019cross}
have shown state-of-the-art results 
in
 unsupervised language translation,
 outperforming 
GPT \cite{radford2018improving}, BERT \cite{devlin2018bert}, 
and other previous methods \cite{lample2017unsupervised, lample2018phrase}.

\subsubsection{{Temperature Control}}

Changing distributions by controlling the temperature has been used in Bayeasian learning and sample generation.
\cite{heek2019bayesian} and \cite{wenzel2020good} reported that sampling weights from its cooled posterior distribution
improves the predictive performance in Bayesian learning.
Higher quality images were generated from a reduced-temperature model in
\cite{parmar2018image, dahl2017pixel, kingma2018glow}. 
\cite{dahl2017pixel} used a tempered softmax for super resolution. 
In contrast to previous works that cool down 
estimated distributions (Bayes posterior or predictive distributions),
our approach  
cools down the input test distribution to make fringe samples more typical
for unsupervised domain translation.


\section{Cooling Down Test Distributions}
\label{sec:proposed_method}

Our proposed method relies on two basic tools, the Metropolis-adjusted Langevin algorithm and a denoising autoencoder.
After introducing those basic tools, we describe our method and its extensions.


\subsection{{Metropolis-adjusted Langevin Algorithm}}

The Metropolis-adjusted Langevin algorithm (MALA)
is an efficient Markov chain Monte Carlo (MCMC) sampling method
that uses the gradient of the energy (negative log-probability $E(\bfx) = - \log p(\bfx)$). 
Sampling is performed sequentially by
\begin{align}
\bfx_{t+1} = \bfx_{t} + \alpha \bfnabla_{\bfx} \log p(\bfx_{t}) + \bfnu,
\label{eq:MalaDAE}
\end{align}
where $\alpha$ is the step size, and
$\bfnu$ is a random perturbation subject to $\mathcal{N}(\bfzero,\delta^{2}\bfI_L)$.
By appropriately controlling the step size $\alpha$ and the noise variance $\delta^{2}$,
the sequence is known to converge to the distribution $p(\bfx)$.%
\footnote{
For convergence, a rejection step after applying Eq.\eqref{eq:MalaDAE} is required.
However, it was observed that a variant, called MALA-approx \cite{Nguyen}, without the rejection step gives reasonable sequence for moderate step sizes.
We use MALA-approx in our proposed method.
}
\cite{Nguyen} successfully generated high-resolution, realistic,
and diverse artificial images
by MALA.



    

\subsection{{Denoising Autoencoders (DAE)}}
\label{secDAE}

A denoising autoencoder (DAE) \cite{Vincent08,Bengio13} is trained so that data samples contaminated with artificial noise are cleaned.  
Specifically, (an estimator) for the following reconstruction error is minimized:
\begin{align}
\Obj (\bfr) &=\expect{ \|\bfr (\bfx + \bfepsilon) -  \bfx  \|^2  }{p(\bfx) p(\bfepsilon)},
\label{eq:DAEObjective}
\end{align}
where $\expect{ \cdot}{p}$ denotes the expectation over the distribution $p$,
$\mathbb{R}^L \ni \bfx  \sim p(\bfx)$ is a data sample, and $\bfepsilon \sim p(\bfepsilon) = \mathcal{N}_L (\bfzero, \sigma^2 \bfI)$ is artificial Gaussian noise with mean zero and variance $\sigma^2$.
\cite{Alain14} discussed the relation between DAEs and contractive autoencoders, and proved the following useful property of DAEs:
\begin{proposition}\cite{Alain14}
\label{prpt:DAEResidual}
Under the assumption that $\bfr(\bfx) = \bfx + o(1)$,
the minimizer of the DAE objective Eq.\eqref{eq:DAEObjective} satisfies
\begin{align}
\bfr(\bfx) - \bfx  = \sigma^2 \bfnabla_{\bfx} \log p(\bfx) 
+ o(\sigma^2),
\label{eq:ResidualScore}
\end{align}
as $\sigma^2 \to 0$.
\end{proposition}
Proposition~\ref{prpt:DAEResidual} states that 
a DAE trained with a small $\sigma^2$ can be used to estimate the gradient of the log probability, i.e.,
\begin{align}
\bfnabla_{\bfx} \log p(\bfx) 
\approx
\widehat{\bfg}(\bfx) \equiv \frac{ \bfr(\bfx) - \bfx}{\sigma^2}.
\label{eq:DAEEstimator}
\end{align}





\subsection{Langevin Cooling (\method{})}
\label{sec:cooling_math}



As discussed in Section~\ref{sec:Inroduction},
we hypothesize that domain translation (DT) methods can work poorly
on test samples lying at the \emph{fringe} of the data distribution.
We therefore propose to drive such fringe samples towards the high density area, where the DNN is better trained.
Specifically, 
we apply MALA Eq.\eqref{eq:MalaDAE} to each test sample with the step size $\alpha$ and the variance of the random perturbation satisfying
the following inequality:
\begin{align}
2 \alpha > 
 \delta^2   .
\label{eq:CoolDownCondition}
\end{align}
If $2\alpha = \delta^2$,
MALA can be seen as a discrete approximation to the (continuous) Langevin dynamics,
\begin{align}
\frac{ d \bfx}{d t}
&=  \bfnabla_{\bfx} \log p(\bfx) + \sqrt{2}\frac{ d \bfW}{d t},
\label{eq:ContinuousBrownianMotion}
\end{align}
where $\bfW$ is the Brownian motion.
The dynamics Eq.\eqref{eq:ContinuousBrownianMotion} is known to converge to $p(\bfx)$
as the equilibrium distribution
\cite{Roberts96,Roberts98}.
By setting the step size and the perturbation variance so that Inequality \eqref{eq:CoolDownCondition} holds,
we can approximately draw samples from the distribution with \emph{lower temperature}, as shown below.

By seeing the negative log probability as the energy $E(\bfx) =- \log p(\bfx)$, we can see $p(\bfx)$  as the Boltzmann distribution with the inverse temperature equal to $\beta = 1$:
\begin{align}
p_\beta(\bfx) & = \frac{1}{Z_\beta} \exp \left(- \beta E(\bfx) \right),
\label{eq:PBeta}
\end{align}
where $Z_\beta = \int \exp \left(- \beta E(\bfx) \right) d\bfx$ is the partition function.
The following theorem holds:
\begin{theorem}
\label{thrm:TemperatureChange}
In the limit where $\alpha, \delta^2 \to 0$ with their ratio $\alpha/ \delta^2$ kept constant,
the sequence of MALA Eq.\eqref{eq:MalaDAE} converges to $p_\beta(\bfx)$
for 
\begin{align}
\beta = \frac{2\alpha}{\delta^2}.
\label{eq:BetaEffective}
\end{align}
\end{theorem}
(Proof)
As $\alpha$ and $\delta^2$ go to $0$, MALA Eq.\eqref{eq:MalaDAE} converges to the following dynamics:
\begin{align}
\frac{ d \bfx}{d t}
&=   \bfnabla_{\bfx} \log p(\bfx) + \frac{\delta}{\sqrt{\alpha}} \frac{ d \bfW}{d t},
\notag
\end{align}
which is equivalent to 
\begin{align}
\frac{ d \bfx}{d t}
&=\frac{2 \alpha}{\delta^2}  \bfnabla_{\bfx} \log p(\bfx) + \sqrt{2} \frac{ d \bfW}{d t}.
\label{eq:EquivalentDynamics}
\end{align}
Eq.\eqref{eq:EquivalentDynamics} 
can be rewritten with the Boltzmann distribution
Eq.\eqref{eq:PBeta} with the inverse temperature specified by Eq.\eqref{eq:BetaEffective}:
\begin{align}
\frac{ d \bfx}{d t}
&=  \bfnabla_{\bfx} \log p_\beta(\bfx) + \sqrt{2} \frac{ d \bfW}{d t}.
\notag
\end{align}
Comparing it with Eq.\eqref{eq:ContinuousBrownianMotion},
we find that this dynamics converges to the equilibrium distribution $p_\beta(\bfx) $.
\QED

Theorem~\ref{thrm:TemperatureChange} states that
the ratio between $\alpha$ and $\delta^2$ effectively controls the temperature.
Specifically, we can see MALA Eq.\eqref{eq:MalaDAE} 
as a discrete approximation to the Langevin dynamics converging to the
distribution given by
\begin{align}
p_{2 \alpha / \delta^2}(\bfx) = \frac{p^{2 \alpha / \delta^2}(\bfx)}{ \int p^{2 \alpha / \delta^2}(\bfx)d\bfx },
\notag
\end{align}
of which the probability mass is more concentrated than $p(\bfx)$ if Inequality \eqref{eq:CoolDownCondition} holds.

Our proposed \emph{Langevin cooling} (\method{}) strategy
uses DAE for estimating the gradient,
and
applies MALA for $\beta > 1$
to cool down test samples
before DT is performed.
As illustrated in Figure~\ref{fig:cartoon},
this yields a small move of the test sample towards high density areas in the source domain.
Since the DNN for DT is expected to be well trained on the high density areas, such a small move can result in a significant improvement of the translated image in the target domain, and thus
enhances the DT performance.
We show qualitative and quantitative performance gain by \method{} in the subsequent sections.




\subsection{Extensions}

We can choose two options for \method{}, depending on the application and computational resources.

\subsubsection{Fringe Detection}
\label{sec:fringe_detection}
We can apply fringe detection, in the same way as adversary detection \cite{srinivasan2018counterstrike}.
Namely, assuming that 
the gradient of $\log p(\bfx)$ is large at the fringe of the data distribution,
we identify samples as fringe if 
\begin{align}
\left\| \bfnabla_{\bfx} \log p(\bfx) \right\|_2
 > \xi
 \label{eq:fringe_detection}
\end{align}
for a threshold $\xi > 0$,
and apply MALA only to those samples.
This prevents non-fringe samples already lying high density areas
from being perturbed by Langevin dynamics.

\subsubsection{Gradient Estimation by Cycle}
\label{sec:lcool_cycle}
Another option is to omit to train DAE, and estimate the gradient by a cycle structure that the DNN for DT already possesses.
This idea follows the argument in \cite{Nguyen},
where MALA is successfully used to generate high-resolution, realistic,
and diverse artificial images.
The authors argued that DAE for estimating the gradient can be replaced with any cycle (autoencoding) structure in their application.
    In our image translation experiment, we use CycleGAN as the base method, and therefore, we can estimate 
    the gradient by
    \begin{align}
    \bfnabla_{\bfx} \log p(\bfx) 
    \approx
    \widehat{\bfg}_{\mathrm{Cycle}}(\bfx)
    &\equiv \gamma \left( \bfF(\bfG(\bfx)) - \bfx \right)
    \label{eq:CycleGANEstimator}
    \end{align}
    for some $\gamma > 0$, 
    where $\bfG$ corresponds to the mapping of the CycleGAN from the source domain to the target domain and $\bfF$ to its inversion.
    We call this option \method{}-Cycle, which eliminates the necessity of training DAE.
    However, 
    one should use this option with care:
    we found 
    that \method{}-Cycle tends to exacerbate artifacts created by CycleGAN, which will be discussed in detail in Section~\ref{sec:grad_estimation}.

\begin{figure}[!t]
\centering
\includegraphics[width=0.5\textwidth]{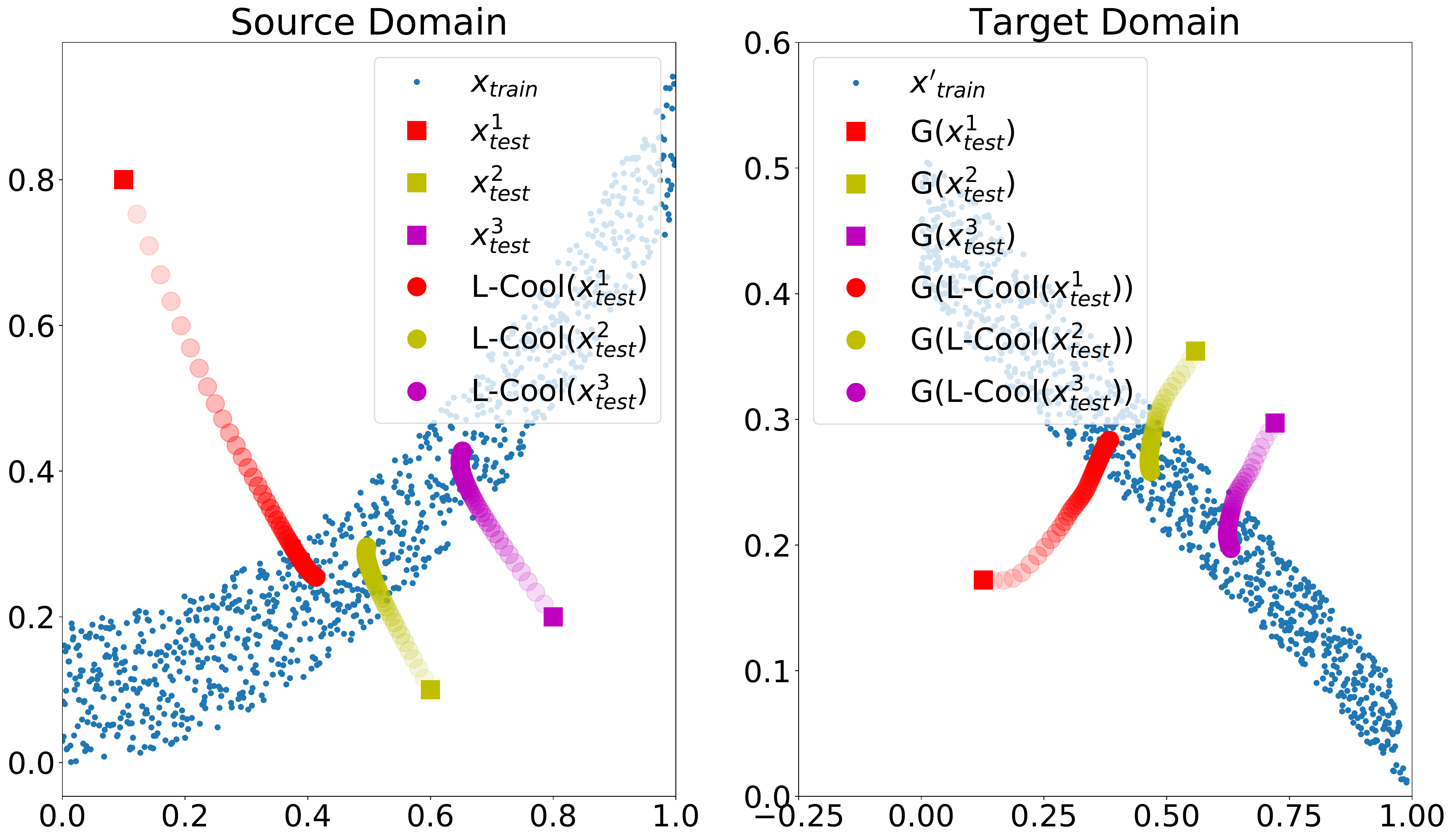} 
\caption{
Toy data demonstration of \method{}, 
which 
drives test samples, 
$x^{1}_{test}, x^{2}_{test}, x^{3}_{test}$, 
towards the data manifold in the source domain (left).
This makes the translated samples 
$\bfG(x^{1}_{test}), \bfG(x^{2}_{test}), \bfG(x^{3}_{test})$ 
by CycleGAN 
more typical
in the target domain (right).
}
\label{fig:toy_data}
\end{figure}

\begin{figure*}[!t]
\centering
\includegraphics[width=\textwidth]{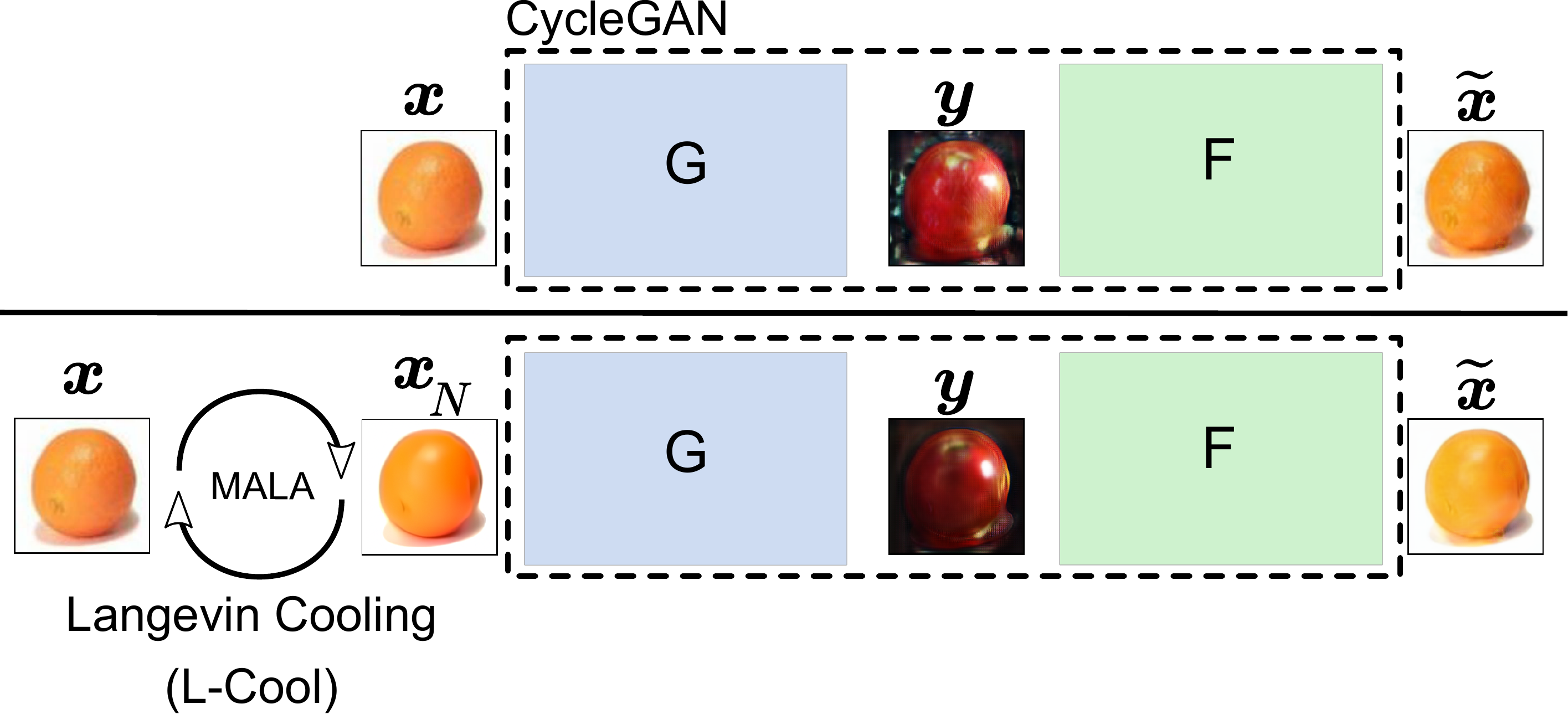}
\caption{
Schematics of (the plain) CycleGAN (top) and \method{} (bottom).  
In CycleGAN,
an encoder, $\bfy  = \bfG(\bfx)$, translates a source sample to a target sample, while a decoder, $\widetilde{\bfx} = \bfF(\bfy)$, translates the target sample back to the source sample. 
In \method{},
a source sample is cooled down by MALA, before being translated by 
CycleGAN.
}
\label{fig:architecture_cyclegan}
\end{figure*}

\section{Demonstration with Toy Data}

We first show the basic behavior of \method{} on toy data.
We generated $1,000$ training samples each in the source and the target domains by
\begin{align}
  \bfx & = (t, 0.75 \times t^2 + \epsilon), 
&  \bfx' & = (t', 0.4 \times t' + \epsilon'),
  \notag
\end{align}
respectively,
where
$t, t' \sim \mathrm{Uniform}(0, 1)$, $\epsilon \sim \mathrm{Uniform}(0, 0.2)$, and $\epsilon' \sim \mathrm{Uniform}(0, 0.1)$. 
Then, a CycleGAN \cite{zhu2017unpaired} with two-layer feed forward networks, $\bfG(\bfx) \to \widehat{\bfx}'$ and $\bfF(\bfx') \to \widehat{\bfx}$, were trained to learn the forward and the inverse mappings between the two domains.
A DAE having 
the same architecture as $\bfG$ with two-layer feed forward network
was also trained on the samples in the source domain.


Blue dots in Figure~\ref{fig:toy_data} show training samples, from which we can see the high density areas 
both in the source (right) and the target (left) domains.
Now we feed three off-manifold test samples $x^{1}_{test}, x^{2}_{test}, x^{3}_{test}$, shown as red, yellow, and magenta squares in the left graph, to the forward (source to target) translator $\bfG$.
As expected, the translated samples $\bfG(x^{1}_{test}), \bfG(x^{2}_{test}), \bfG(x^{3}_{test})$, shown as red, yellow, and magenta squares in the right graph, are not in the high density area (not typical target samples), because $\bfG$ was not trained for those off-manifold samples.
As shown as trails of circles,
\method{} drives the off-manifold samples into the data manifold in the source domain,
which also drives the translated samples into the data manifold in the target domain.
This way, \method{} helps CycleGAN generate typical samples in the target domain by making source samples more typical.

\section{Image Translation Experiments}
\label{sec:ITExperiment}

Next, we demonstrate the performance of \method{} in several image translation tasks.  We use CycleGAN as the base translation method,
and \method{} is performed in the source image space before translation (Figure~\ref{fig:architecture_cyclegan}).


\subsection{ Translation Tasks and Model Architectures}

We used pretrained CycleGAN models, along with the training and the test datasets, publicly available in the
official Github repository\footnote{https://github.com/junyanz/pytorch-CycleGAN-and-pix2pix} of CycleGAN \cite{zhu2017unpaired}. 
Experiments were conducted on the following tasks.
\begin{description}
\item [horse2zebra]
Translation from horse images to zebra images and vice versa.
The training set consists of $1067$ horse images and $1334$ zebra images, subsampled from ImageNet.
Dividing the test set,
we prepared $60$ and $70$ validation images and $60$ and $70$ test images for horse and zebra, respectively.

\item [apple2orange]
Translation from apple images to orange images and vice versa.
The training set consists of $995$ apple images and $1019$ orange images, subsampled from ImageNet.
Dividing the test set,
we prepared $133$ and $133$ validation images and $133$ and $133$ test images for apple and orange, respectively.

\item [sat2map]
Translation from satellite images to map images.
The training set consists of $1096$ satellite images and $1096$ map images, subsampled from Google Maps.
$1098$ and $1098$ images each are provided for test.  Dividing the test set,
we prepared $250$ validation images and $848$ test images.
Although CycleGAN was pretrained in the unsupervised setting,
the dataset is actually paired, i.e., the ground truth map image for each satellite image is available,
which allows quantitative evaluation.

\end{description}
For the first two tasks, we also conducted experiments on the inverse tasks, i.e., zebra2horse and orange2apple.
The validation images were used for hyperparameter tuning for \method{} (see Section~\ref{sec:temp_analysis_image}).  

The CycleGAN model consists of a forward mapping $\bfG$ and a reverse mapping $\bfF$. 
Both $\bfG$ and $\bfF$ have the same architecture including
$2$ downsampling layers followed by $9$ resnet generator blocks and $2$ upsampling layers. 
Each resnet generator block consists of convolution, 
batch normalization \cite{ioffe2015batch} and ReLU layers with residual connections added between every block.

For DAE, we adapted a Tiramisu model \cite{jegou2017one} consisting $67$ layers in total.
The PyTorch \cite{pytorch} code for Tiramisu was obtained 
from a publicly 
available GitHub reporsitory\footnote{https://github.com/bfortuner/pytorch\_tiramisu}. 
The Tiramisu consists of $5$ downsampling layers followed by 
a bottleneck layer 
and $5$ upsampling layers. 
Each downsampling as well as upsampling layer
consists of dense blocks with a growth rate of $16$. 
Each dense block consists of batch normalization \cite{ioffe2015batch}, ReLU, and convolution layers with dense connections \cite{huang2017densely}. 
We trained the DAE on the training images in the source domain
for $200$ epochs by the Adam optimizer with the learning rate set to $0.0002$. 

\begin{figure*}
\centering
    \begin{subfigure}[t]{\textwidth}
        \centering
        \includegraphics[width=0.32\columnwidth, height=7cm,keepaspectratio]{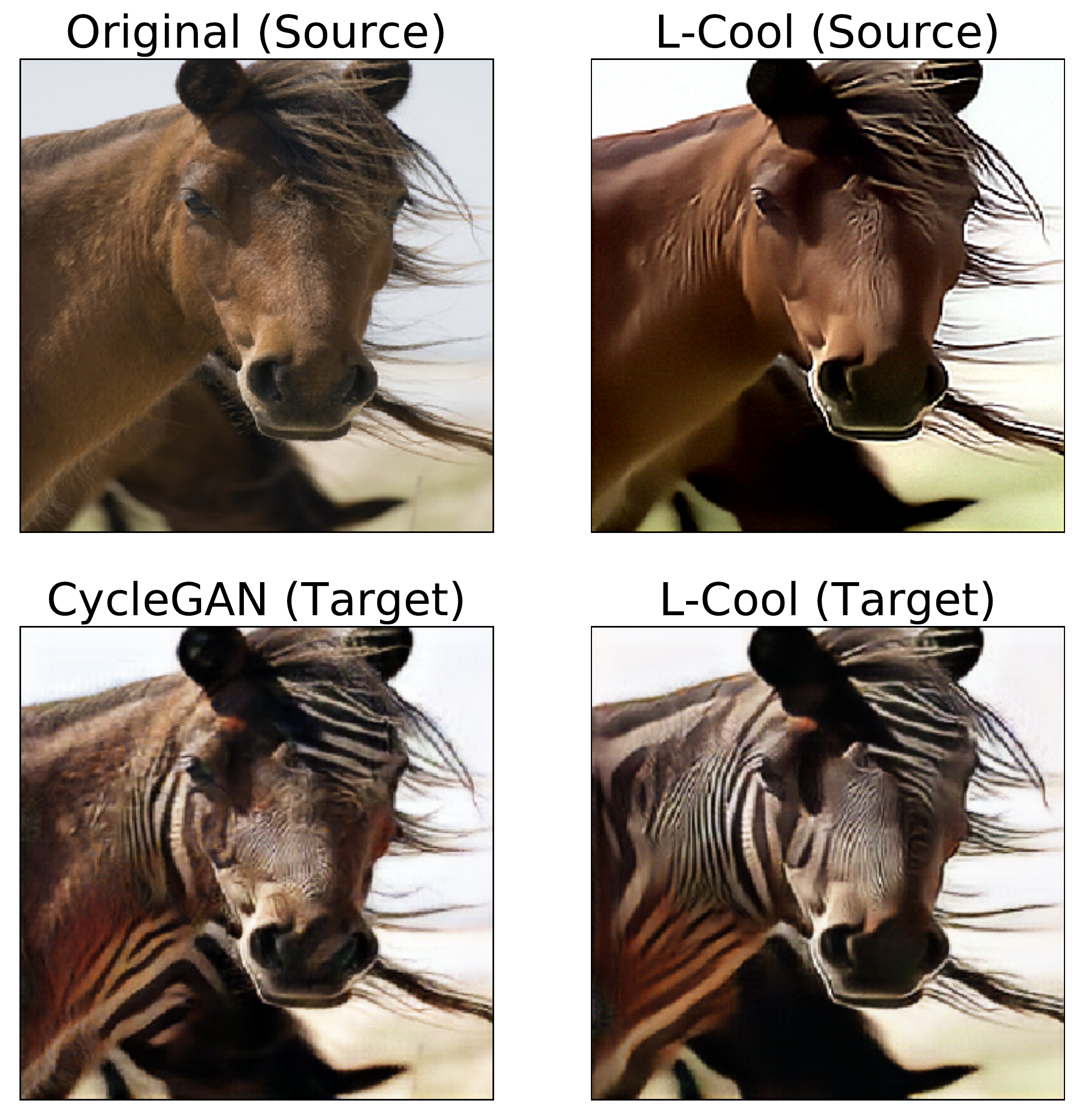}        
        \includegraphics[width=0.32\columnwidth, height=7cm,keepaspectratio]{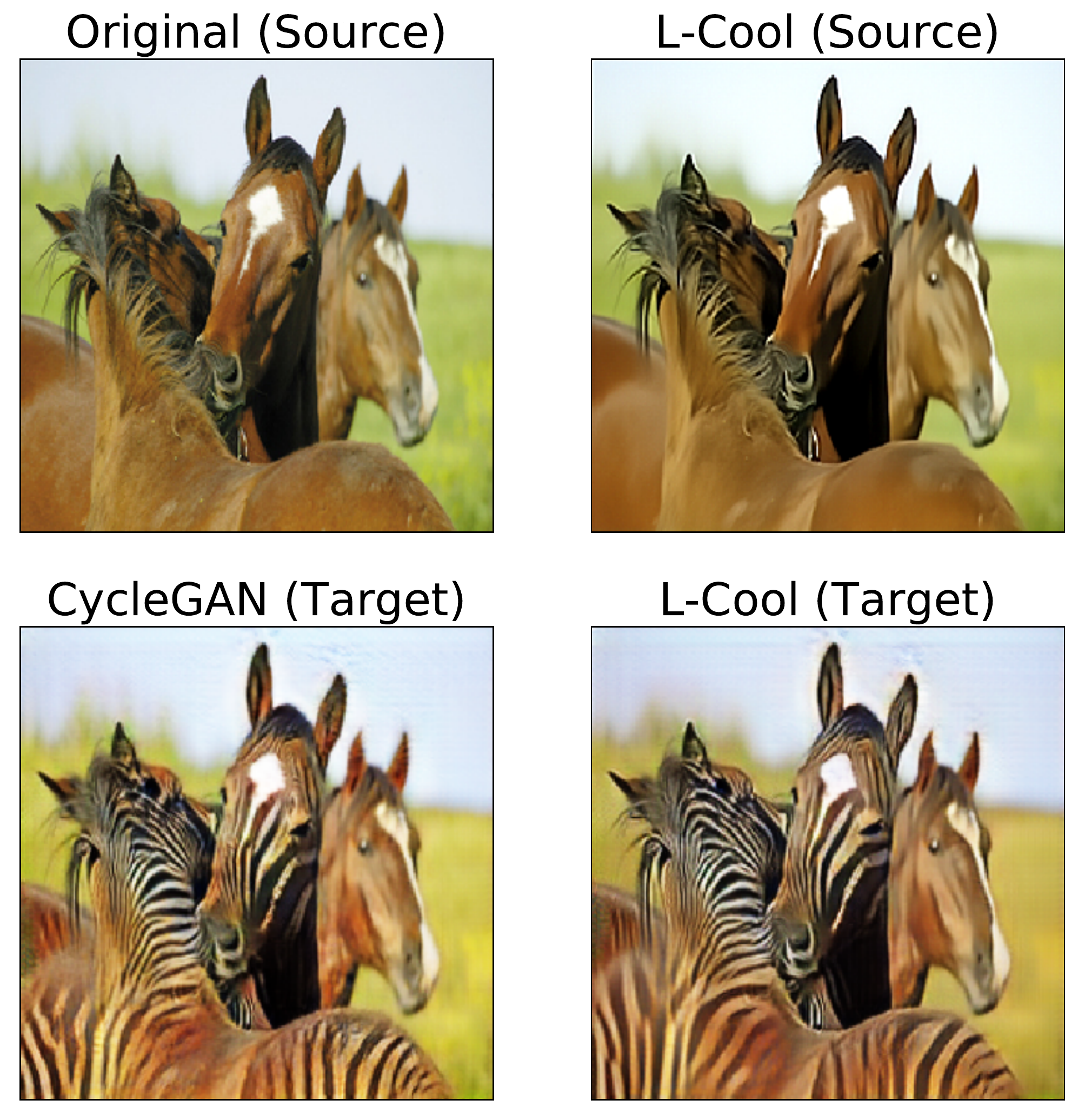}
        \includegraphics[width=0.32\columnwidth, height=7cm,keepaspectratio]{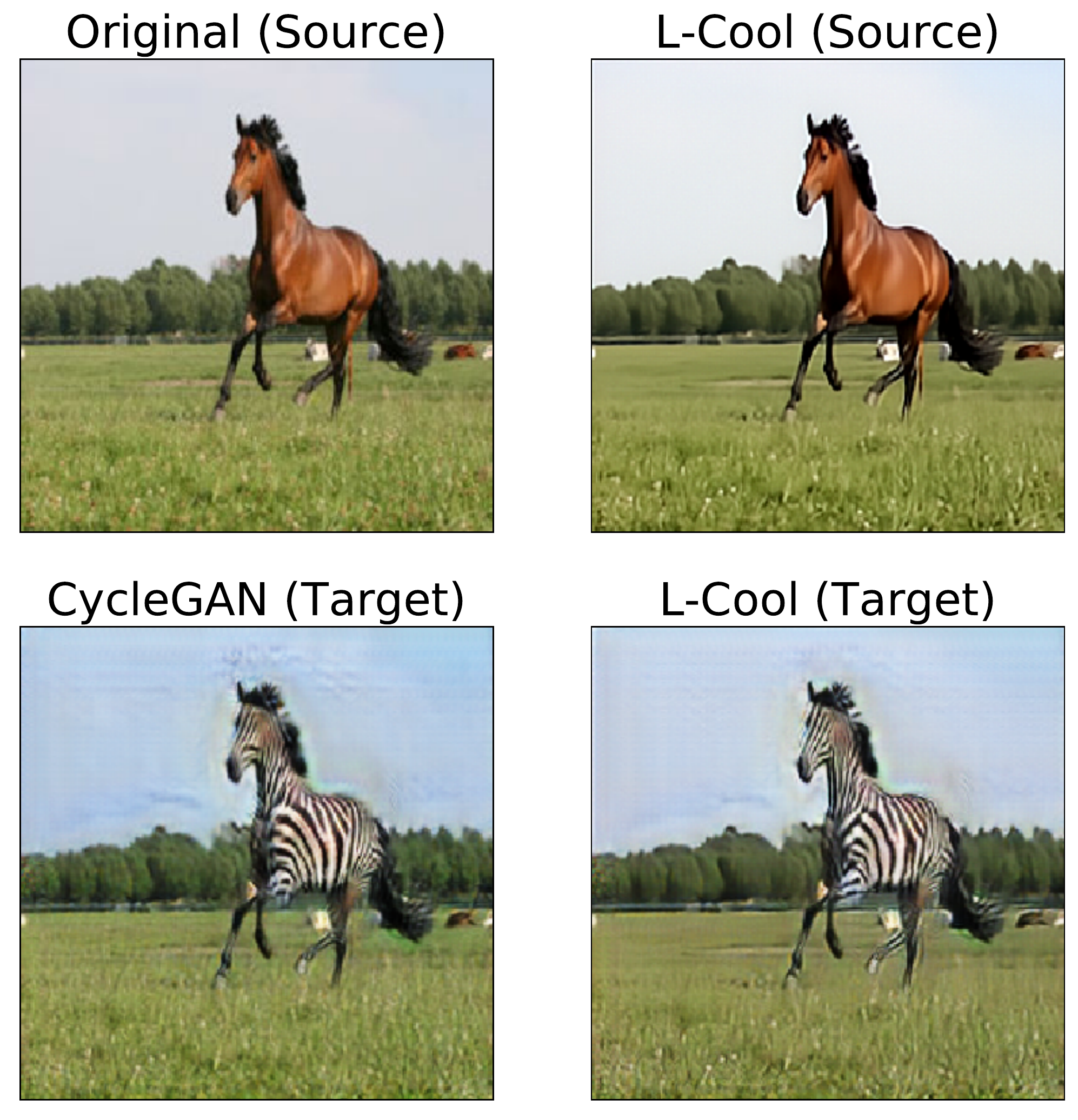}
        \caption{horse2zebra: 
        The contrast of stripes are increased (left and middle) and artifacts around the zebra are reduced (right).
        }           \label{fig:qual_h2z}
    \end{subfigure}
    \begin{subfigure}[t]{\textwidth}
        \centering
        \includegraphics[width=0.32\columnwidth, height=7cm,keepaspectratio]{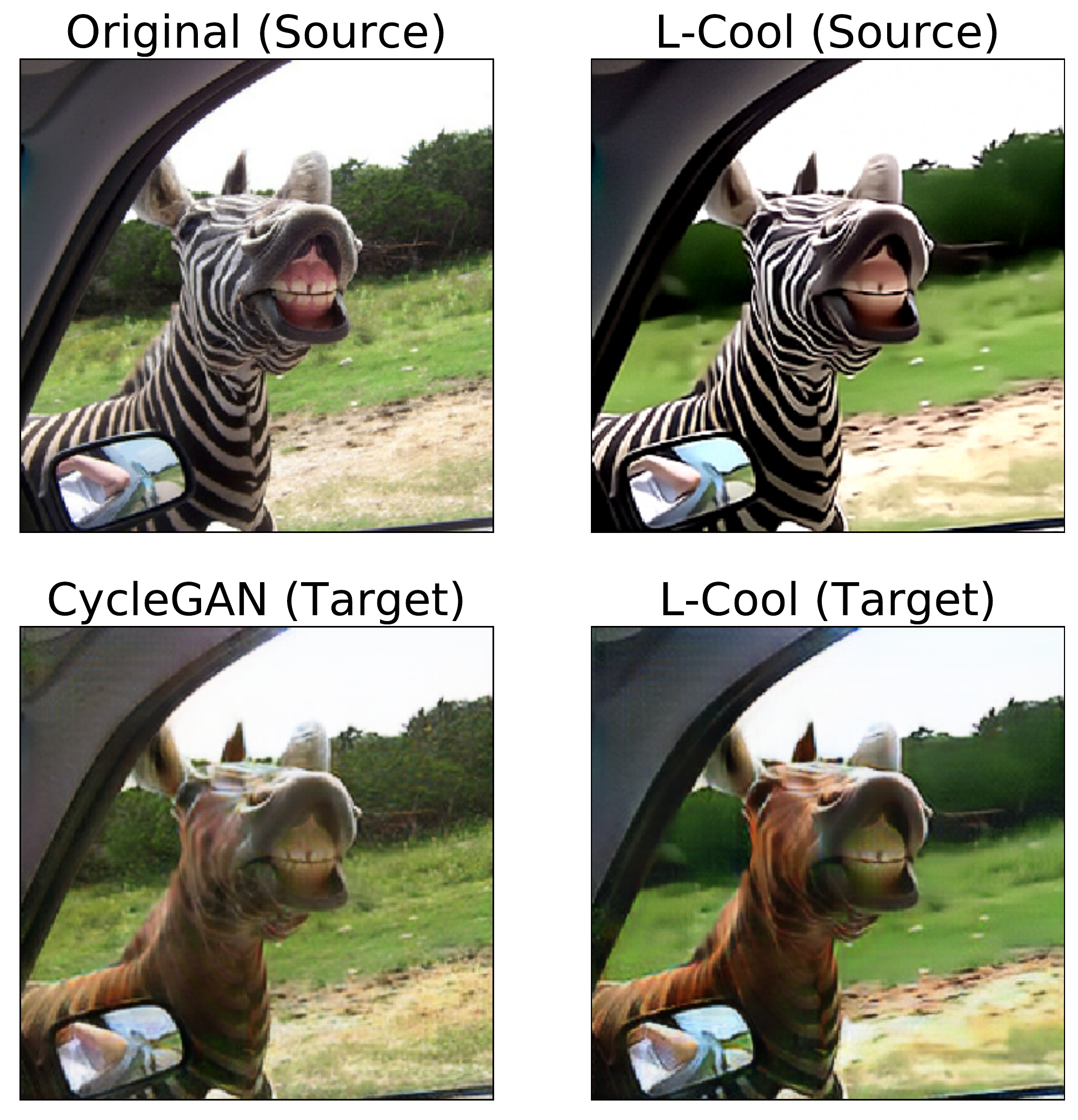}
        \includegraphics[width=0.32\columnwidth, height=7cm,keepaspectratio]{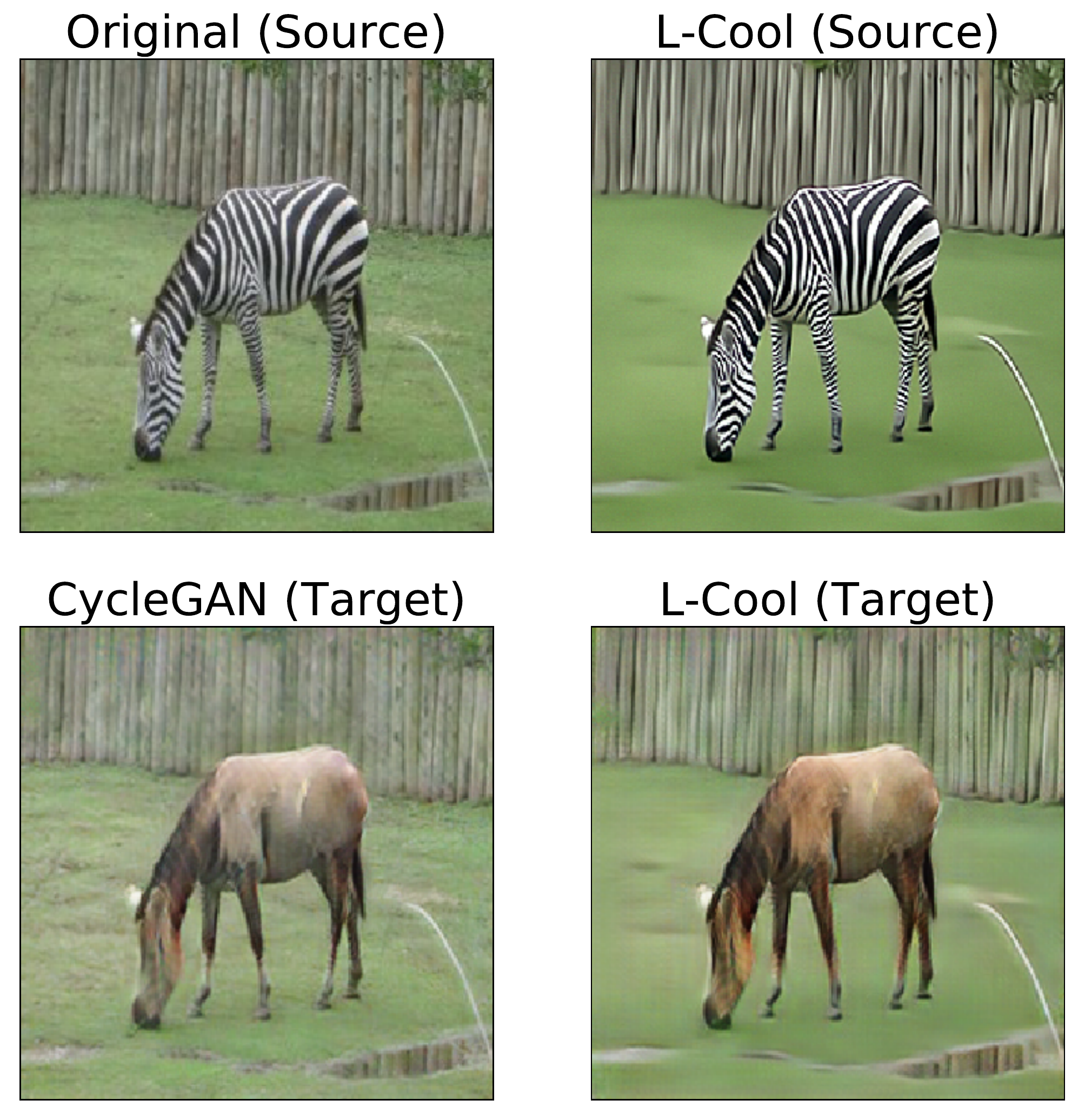}
        \includegraphics[width=0.32\columnwidth, height=7cm,keepaspectratio]{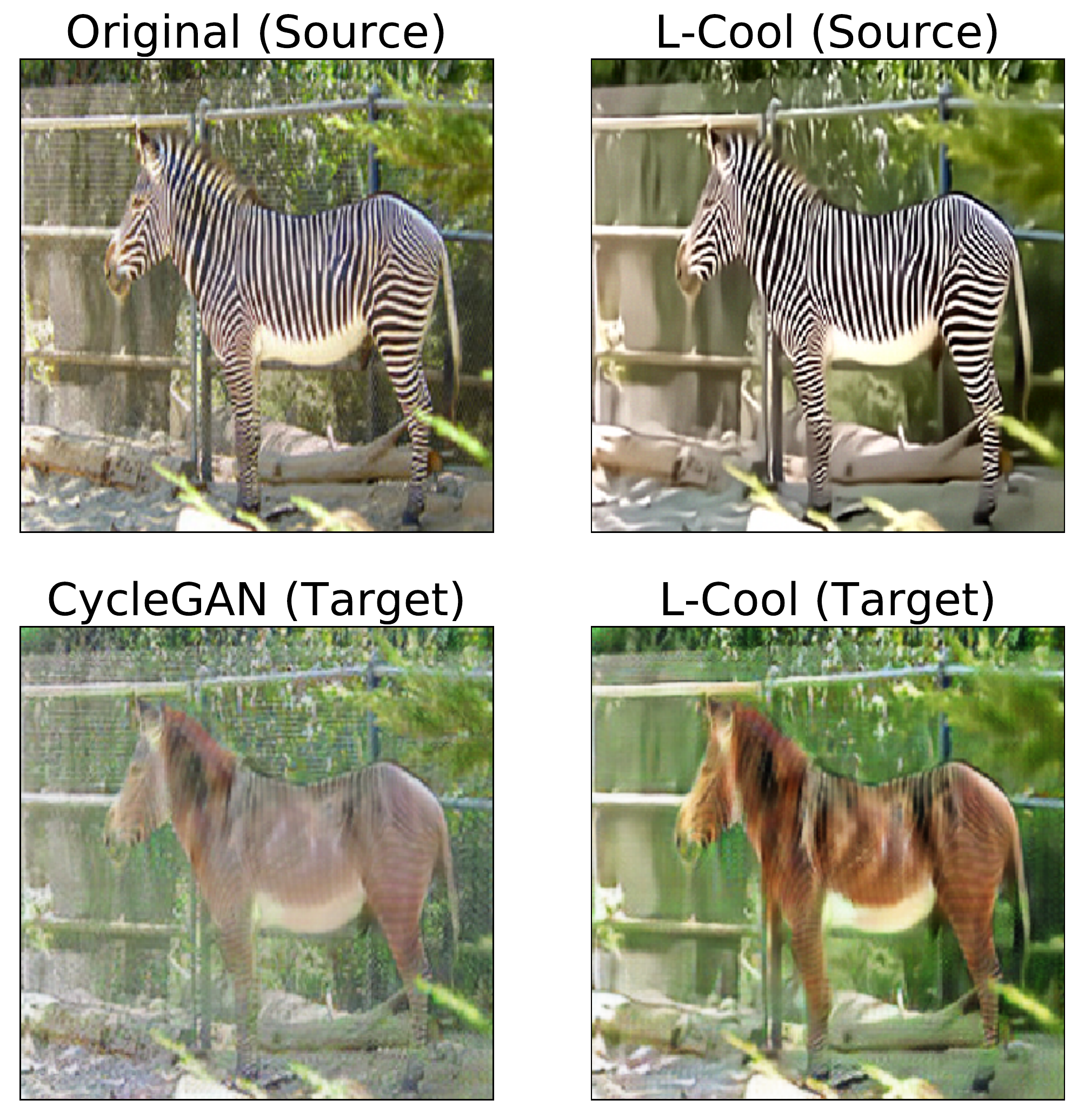}        

        \caption{zebra2horse: 
        The color of the horse body is improved.
        }
        \label{fig:qual_z2h}
    \end{subfigure}
    ~
    \begin{subfigure}[t]{\textwidth}
        \centering
        \includegraphics[width=0.32\columnwidth,height=7cm,keepaspectratio]{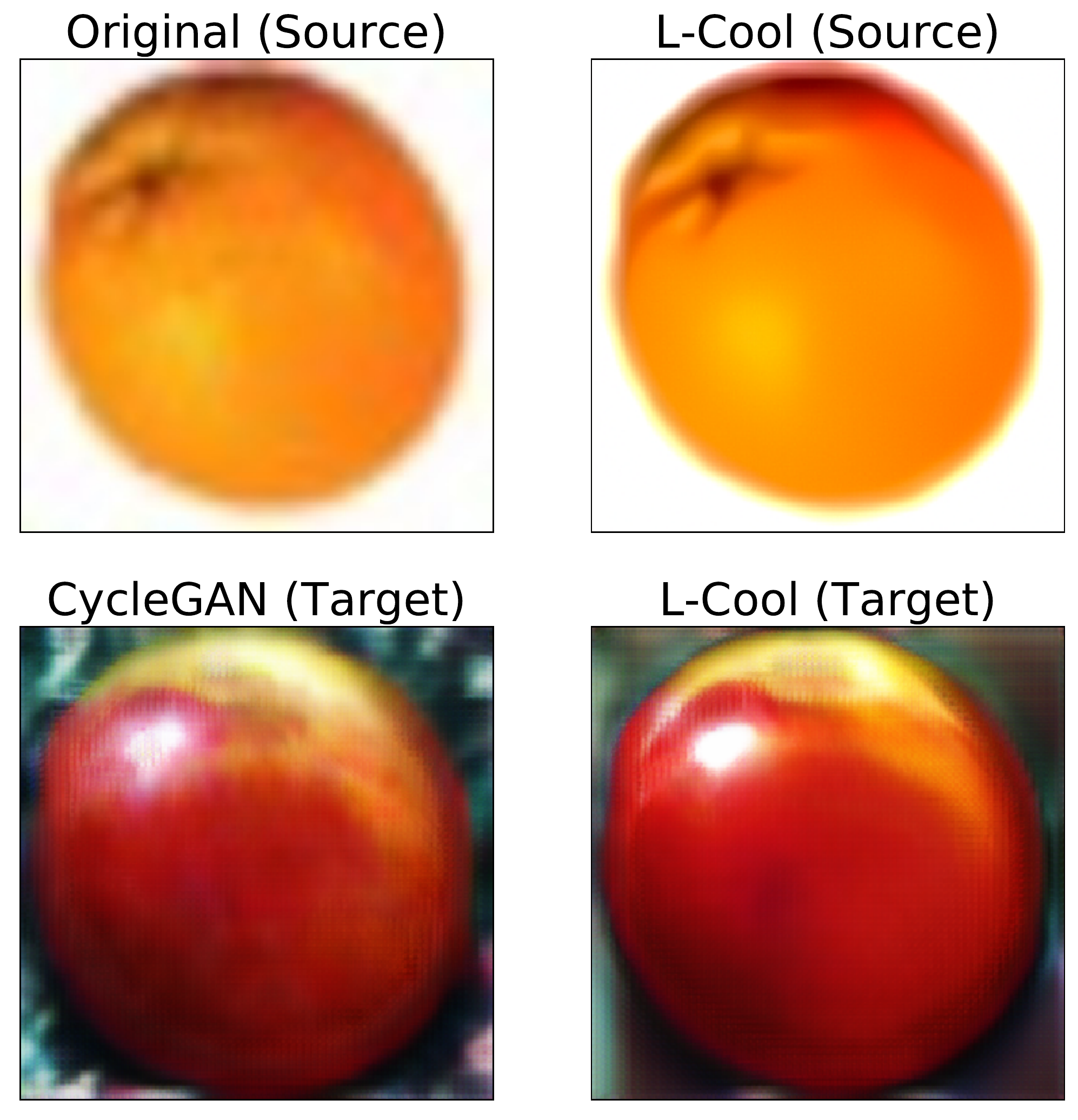}        
        \includegraphics[width=0.32\columnwidth,height=7cm,keepaspectratio]{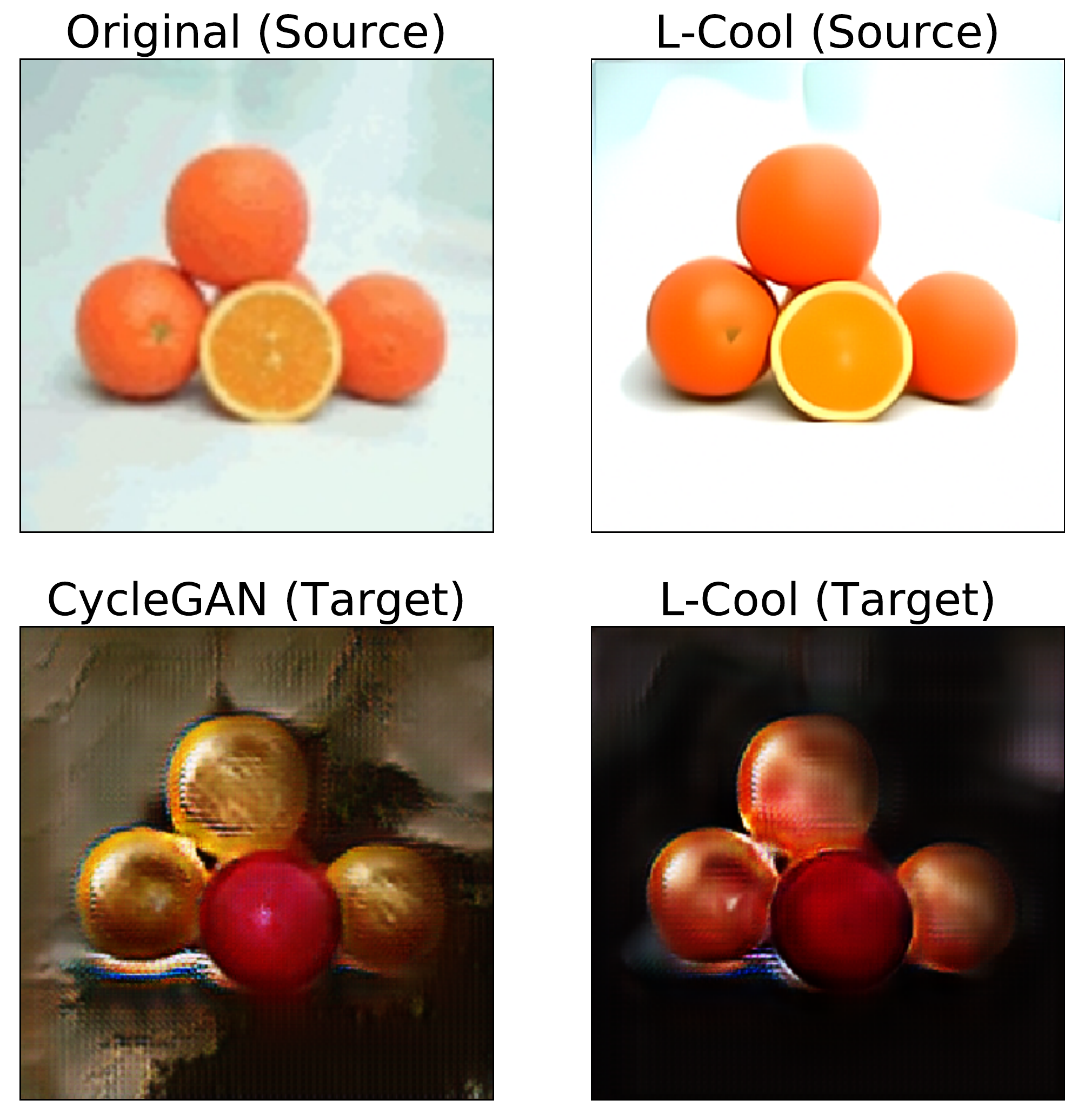}
        \includegraphics[width=0.32\columnwidth,height=7cm,keepaspectratio]{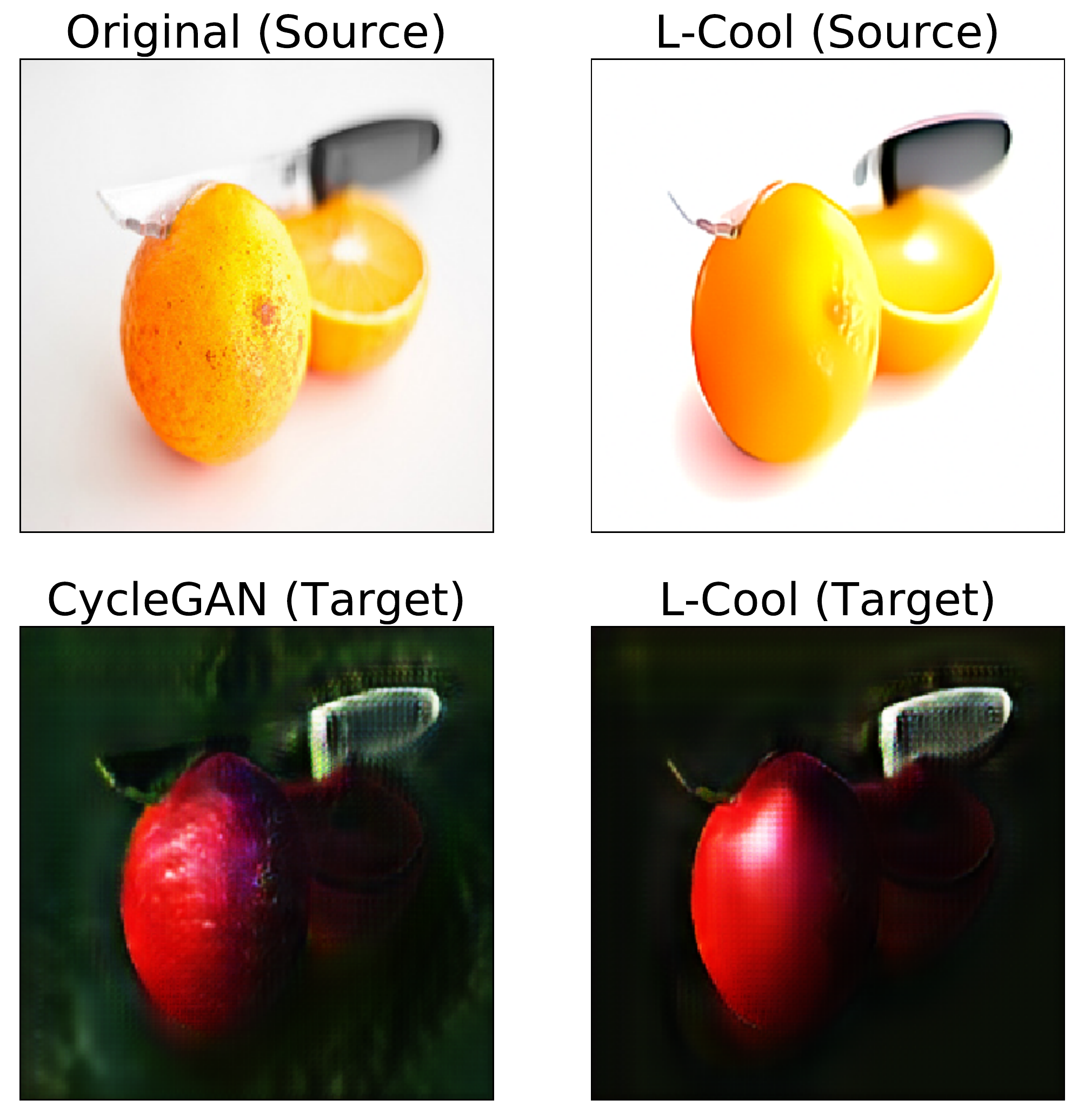}
        \caption{orange2apple: 
        The texture and the color of apples are improved, and artifacts in the background are reduced.
        }
         \label{fig:qual_o2a}
    \end{subfigure}
    \begin{subfigure}[t]{\textwidth}
        \centering
        \includegraphics[width=0.32\columnwidth,height=7cm,keepaspectratio]{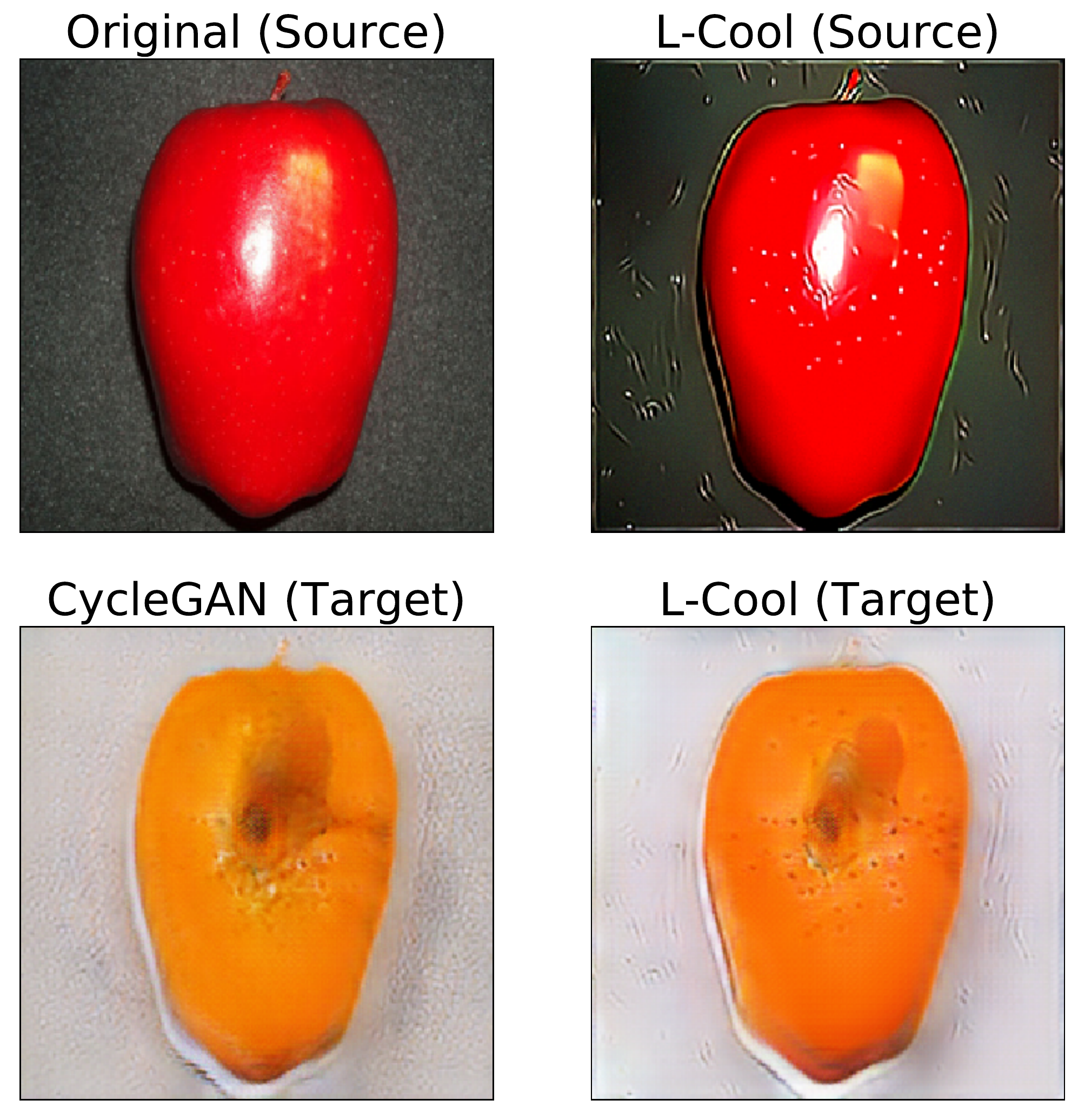}
        \includegraphics[width=0.32\columnwidth,height=7cm,keepaspectratio]{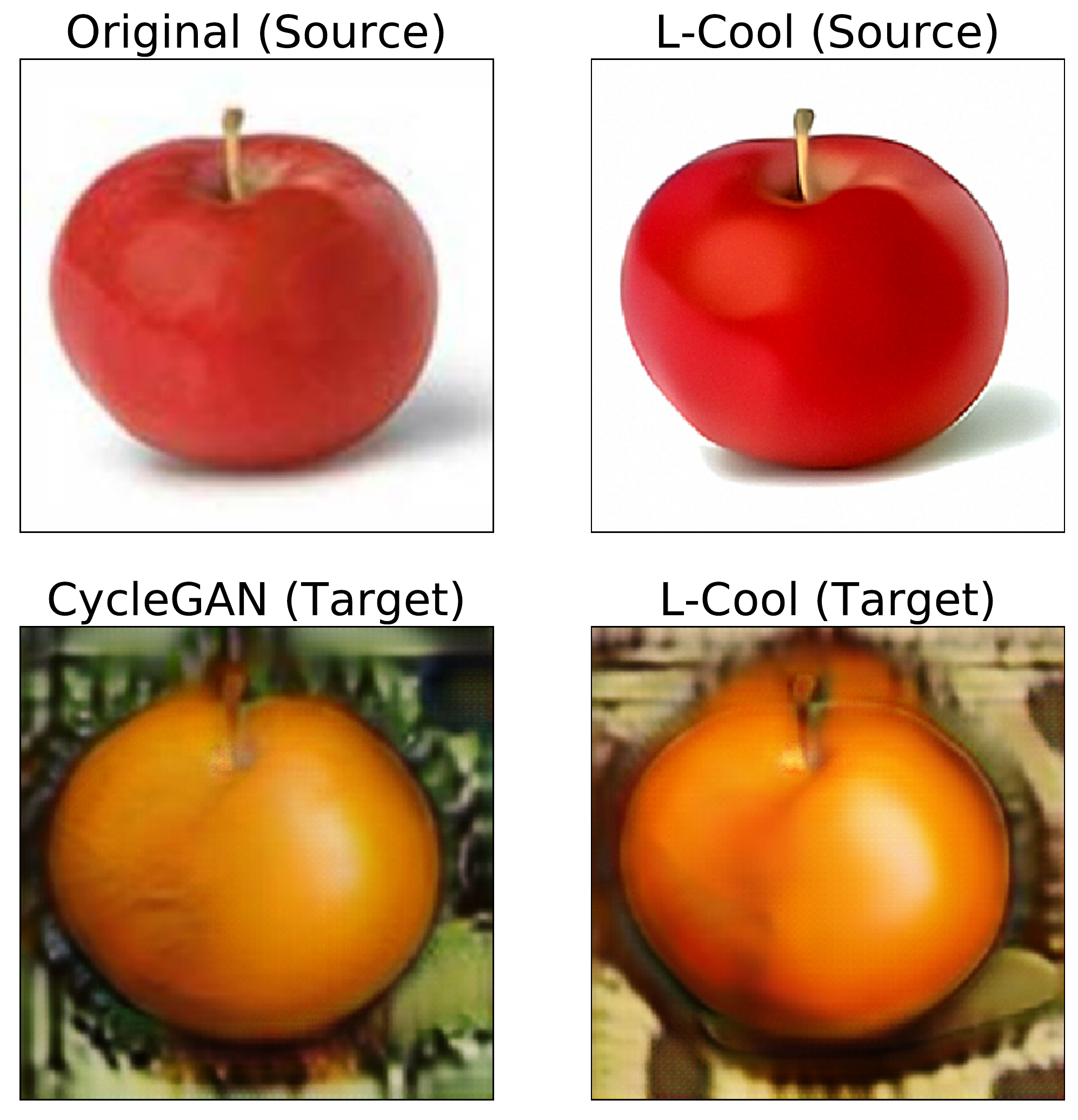}
        \includegraphics[width=0.32\columnwidth,height=7cm,keepaspectratio]{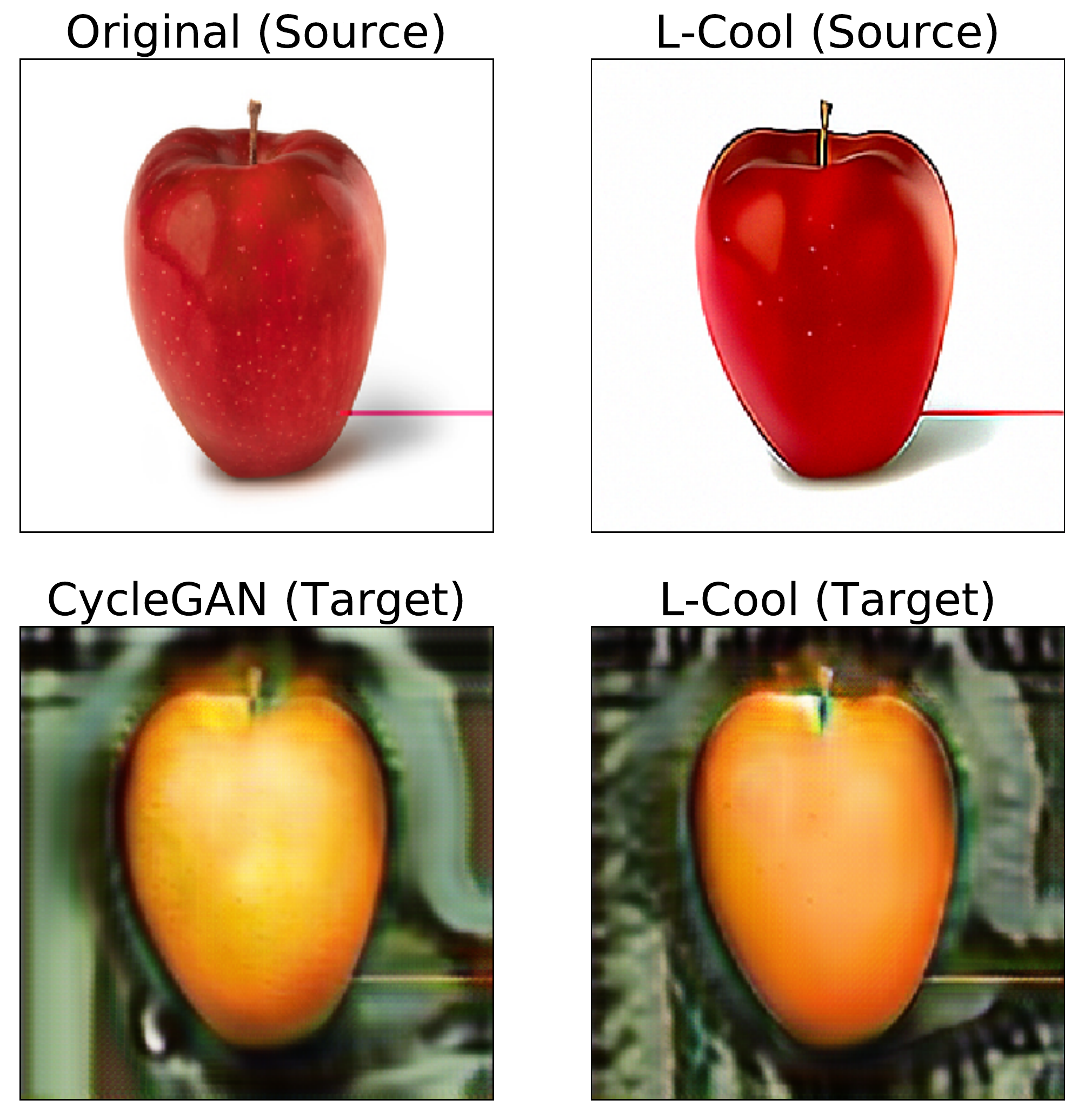}
        \caption{apple2orange: 
        The color of oranges is improved.}
        \label{fig:qual_a2o}
    \end{subfigure}
\caption{
Example results of image translation tasks.
Three examples for each task are shown, and each example shows the original test image (top left) and  
the image after \method{} is applied (top right) in the source domain, and their translated images (bottom left and right) in the target domain. 
}
\label{fig:qualitative_fig}
\end{figure*}

\subsection{Qualitative Evaluation}


  



\begin{figure*}[!t]
\centering

    \begin{subfigure}[t]{0.30\textwidth}
        \centering
        \includegraphics[width=\textwidth]{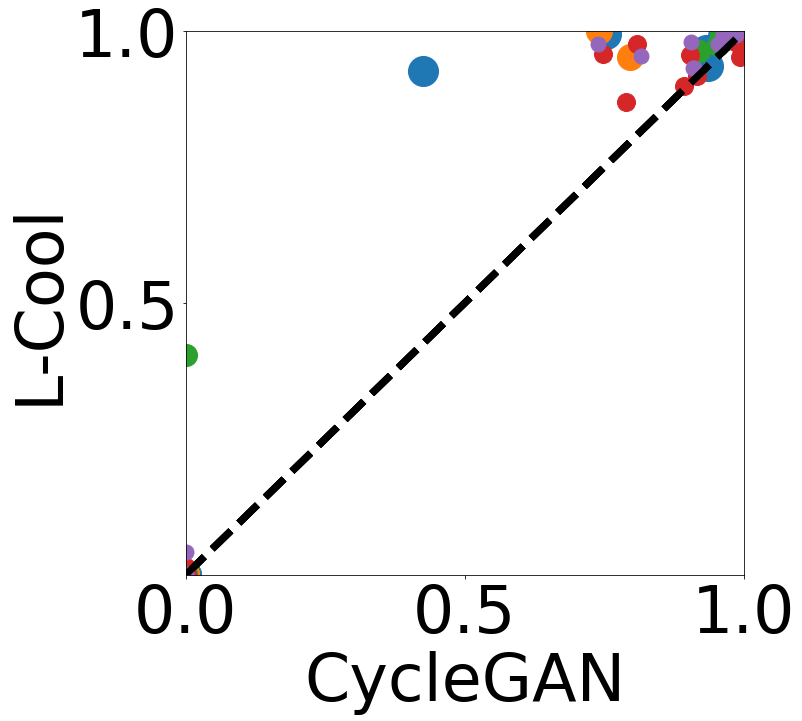}
        \label{fig:probs_classifier_20}
        \caption{\%fringes: $20$}
    \end{subfigure}
    ~
    \begin{subfigure}[t]{0.55\textwidth}
        \centering
        \includegraphics[width=\textwidth]{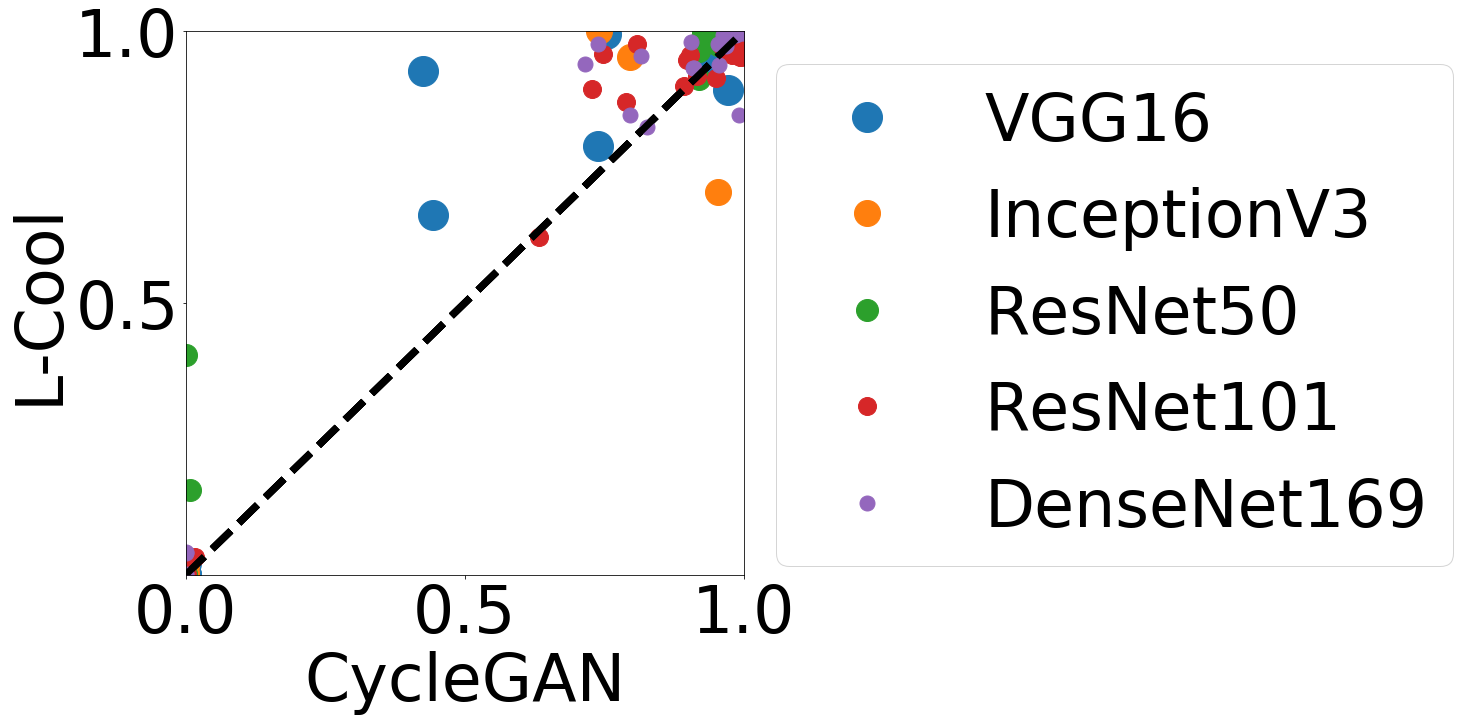}
        \label{fig:probs_classifier_40}
        \caption{\%fringes: $40$}
    \end{subfigure}
    ~
    \\
    \begin{subfigure}[t]{0.30\textwidth}
        \centering
        \includegraphics[width=\textwidth]{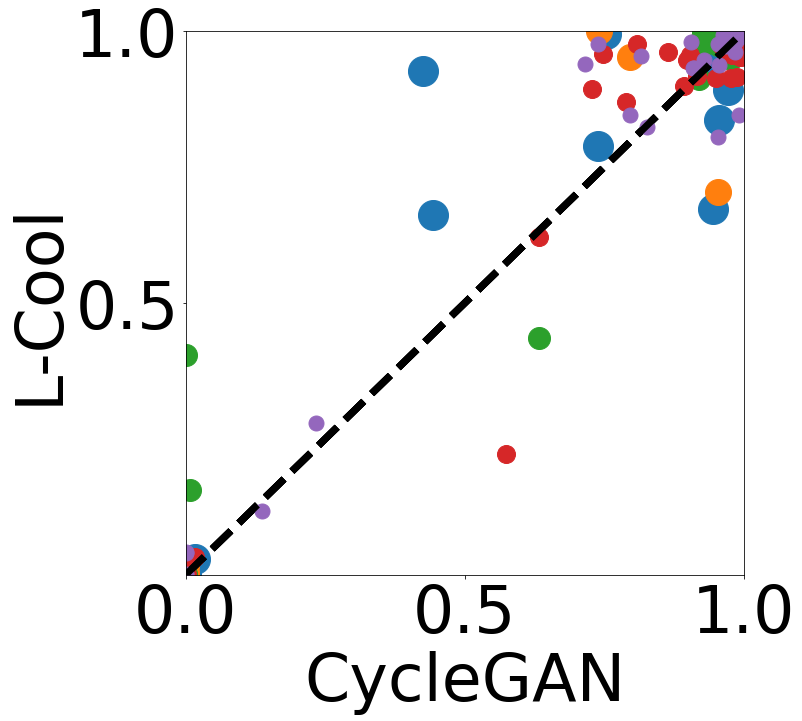}
        \label{fig:probs_classifier_60}
        \caption{\%fringes: $60$}
    \end{subfigure}
    ~
    \begin{subfigure}[t]{0.30\textwidth}
        \centering
        \includegraphics[width=\textwidth]{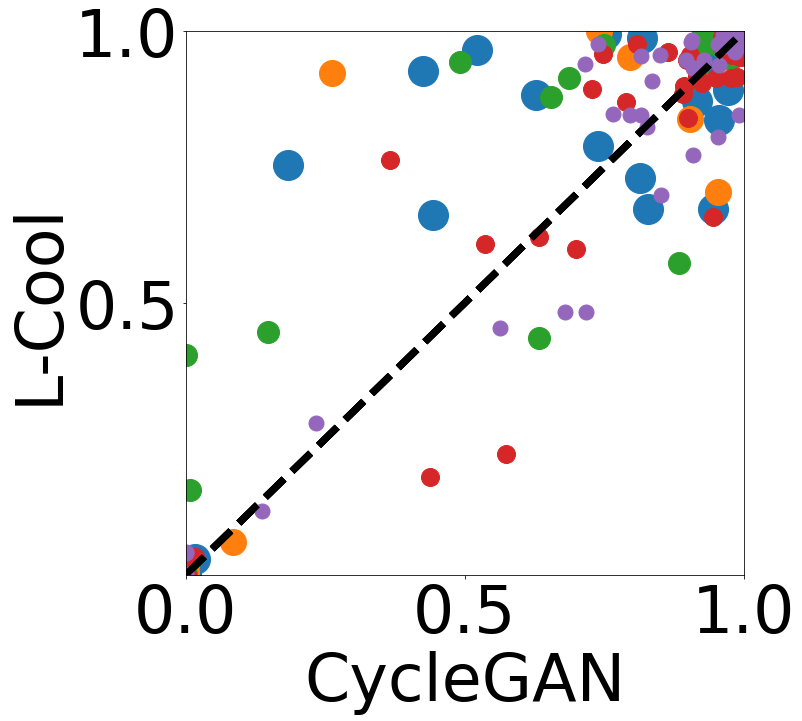}
        \label{fig:probs_classifier_80}
        \caption{\%fringes: $80$}
    \end{subfigure}
    ~
    \begin{subfigure}[t]{0.30\textwidth}
        \centering
        \includegraphics[width=\textwidth]{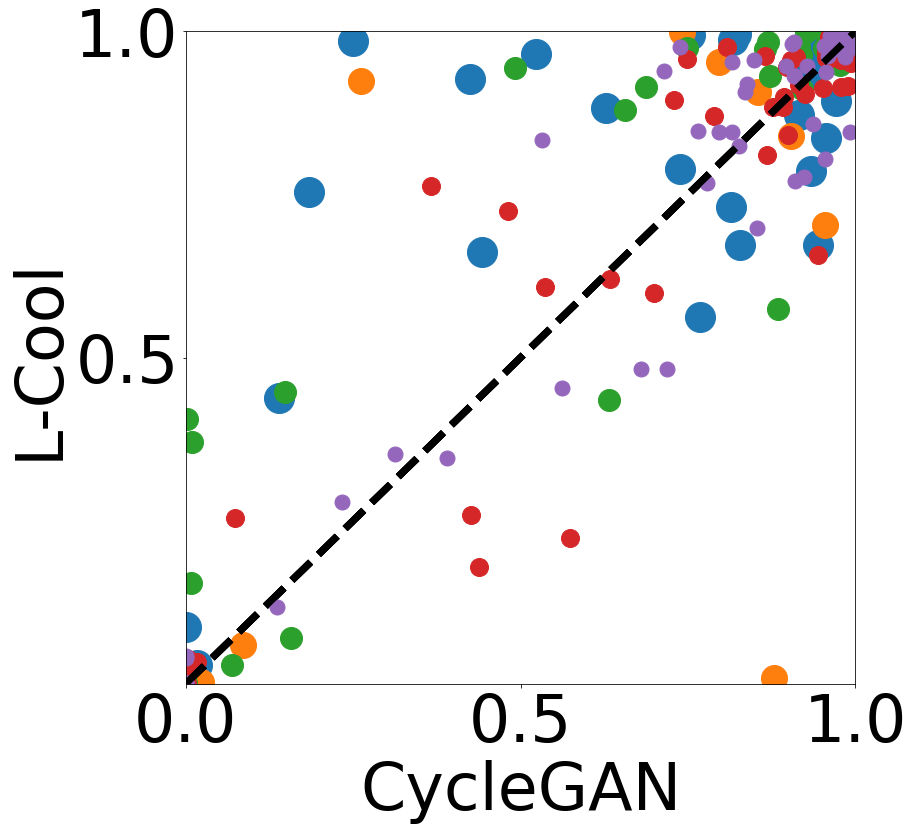}
        \label{fig:probs_classifier_100}
        \caption{\%fringes: $100$}
    \end{subfigure}

\caption{
Likeness to zebra images evaluated by the probability output $p(\bfy=\textrm{zebra}|\bfx)$ of pretrained classifiers for the translated images by CycleGAN (horizontal axis) and by \method{} (vertical axis).
Each panel plots the fringe samples identified by the fringe detector for different proportions.
We can see that,
consistently for all classifiers (shown in different colors), points tend to be above the equal-likeness dashed line, 
 implying improvement by \method{}.
}
\label{fig:probs_classifiers}
\end{figure*}

\begin{table}[t]
  \caption{
Average likeness to zebra images
over the fringe samples and the classifiers (shown in the legend in Figure~\ref{fig:probs_classifiers}).
For each row, the methods that are not significantly outperformed by the other are bold-faced, according to the Wilcoxon signed rank test for $p = 0.05$.
    }
  \label{table:probs_classifiers}
  \centering

  \begin{tabular}{c c c}
    \toprule
    \% fringes & CycleGAN & \method{} 
    \\

\midrule
\begin{tabular}{@{}c@{}}20 \\ 40 \\ 60 \\ 80\\100\end{tabular} 
    & \begin{tabular}{@{}c@{}}0.6910 \\ 0.7872 \\ \textbf{0.8023 }\\ \textbf{0.8138}\\\textbf{0.8022}\end{tabular} 
    & \begin{tabular}{@{}c@{}}\textbf{0.7385}\\ \textbf{0.8145} \\ \textbf{0.8167} \\ \textbf{0.8331} \\ \textbf{0.8211}\end{tabular}  \\
    \bottomrule
\end{tabular}
\end{table}

Figure~\ref{fig:qualitative_fig} shows some example results of horse2zebra, zebra2horse, apple2orange, and orange2apple tasks.
We see that \method{} moves original source images more typical (in terms of color and smoothness), which results in improved translated images, e.g., 
more stripes in (a) horse2zebra, 
more brown color in the horse body in (b) zebra2horse, 
better texture and color in (c) apple2orange and 
(d) orange2apple,
and removal of artifacts in general.

\subsection{Quantitative Evaluation}
\label{sec:Image.Quantitative.Evaluation}

In order to
confirm that \method{} generally improves the image translation performance,
we conducted two experiments that quantitatively evaluate the performance.

\subsubsection{Likeness Evaluation by Pretrained Classifiers}
\label{sec:LikenessEvaluation}

Focusing on horse2zebra, we evaluated the likeness of the translated images
to zebra images by using state-of-the-art classifiers,
including
VGG16 \cite{vgg}, 
InceptionV3 \cite{szegedy2016rethinking}, Resnet50 \cite{xie2017aggregated}, Resnet101 \cite{zagoruyko2016wide}, and DenseNet169 \cite{huang2017densely} 
pretrained on the ImageNet dataset \cite{imagenet_cvpr09}.
Specifically, we evaluated and compared the probability outputs (i.e., after soft-max) of the classifiers for the translated images by plain CycleGAN and those by \method{}.
We applied fringe detection, Eq.\eqref{eq:fringe_detection}, with the threshold $\xi$ adjusted so that specified proportions (20\%, 40\%, 60\%, 80\%, and 100\%) of the test samples are identified as fringe. 
Note that $100\%$ fringe samples correspond to \method{} without fringe detection (all test samples are cooled down by MALA).

Figure~\ref{fig:probs_classifiers} shows scatter plots of 
likeness to zebra images, i.e., the probability $p(\bfy=\textrm{zebra}|\bfx)$ evaluated by pretrained classifiers.
The five panels respectively plot the
$20,40,60$, $80$ and $100 \%$ fringe samples.
In each plot, the horizontal axis corresponds to the likeness of the transferred images by CycleGAN, while the vertical axis corresponds to the likeness of the transferred images by \method{}.  The dashed line indicates the equal-probability, i.e., the points above the dashed line imply the improvement by \method{}.

We observe that all classifiers tend to give higher probability to 
the images translated after \method{} is applied.  
We emphasize that \method{} uses no information on the target domain---DAE is trained purely on the samples in the source domain, and MALA drives samples towards high density areas in the source domain, 
independently from the translation task.
The hyperparameters for the Langevin dynamics were set to $\alpha = 0.005$, $\beta^{-1} = 0.001$ and $N = 40$,
which were found optimal on the validation set (see Section~\ref{sec:temp_analysis_image}).
Table~\ref{table:probs_classifiers} shows the average likeness over the fringe samples and the five classifiers.

We observe in Table~\ref{table:probs_classifiers} that,
for smaller proportions of fringe samples (first column), the performance of the plain CycleGAN (second column) is worse, and the performance gain, i.e., the differences between \method{} (third column) and CycleGAN, is larger.
These observations empirically support our hypothesis that CycleGAN does not perform well on fringe samples, and cooling down those samples can improve the translation performance.

\subsubsection{Evaluation on Paired-data}

As mentioned in Section~\ref{sec:TranslationTasks}, sat2map dataset
consists of pairs of satellite images and the corresponding map images,
and therefore
allows us to directly evaluate image translation performance.  
We applied the pretrained CycleGAN to the test satellite images with and without \method{},
and compared the transferred map images with the corresponding ground-truth map images.
Following the evaluation procedure in \cite{liu2017unsupervised},
we counted pixels as \emph{correct} if the color mismatch (i.e., the Euclidean distance between the transferred map and the ground-truth map in the RGB color space) is below 16.

\begin{table}[!t]
  \caption{Average pixel-wise accuracy in the sat2map task. For each row, the methods that are not significantly outperformed by the other are bold-faced, according to the Wilcoxon signed rank test for $p = 0.05$.}
\label{table:sat2maps}
  \centering

  \begin{tabular}{ c c c  c }
    \toprule
    \%fringes & CycleGAN &  \method{} 
    \\
    \midrule
    20    & 61.83& \textbf{62.76} 
    \\ 
    40  &  65.95 & \textbf{66.37} 
    \\   
    60    & 66.37 & \textbf{67.54} 
    \\ 
    80    & 68.56 &  \textbf{68.76} 
    \\ 
    100    & 68.83 & \textbf{69.05} 
    \\ 
    \bottomrule
  \end{tabular}
\end{table}

\begin{figure}[!t]
\centering

\includegraphics[width=0.9\columnwidth]{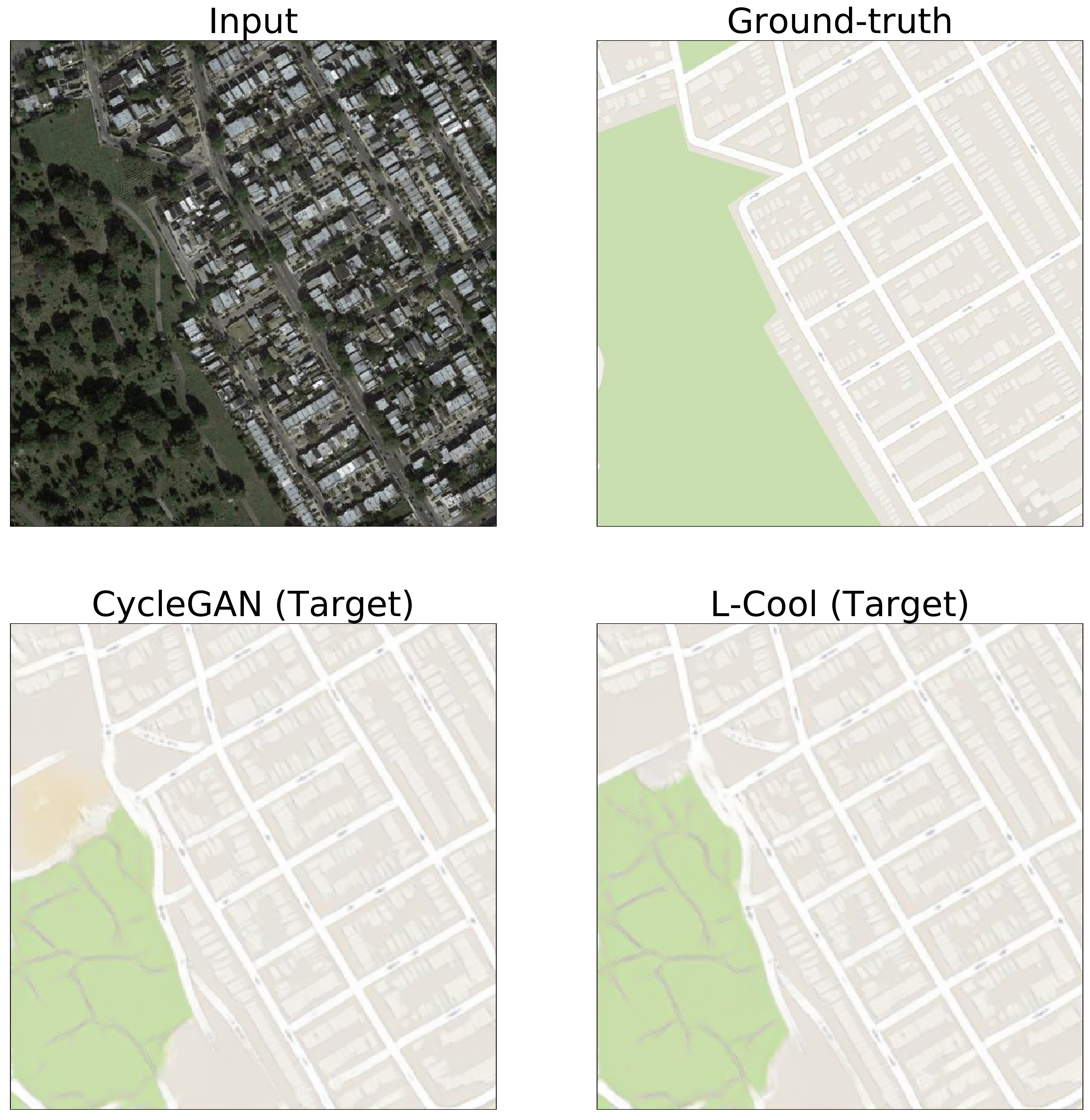}
\caption{
An example of sat2map image translation result.
\method{} result (bottom right) is closer to the ground truth (top right) than the plain CycleGAN (bottom left). 
}
\label{fig:sat2maps}
\end{figure}

\begin{figure*}[t]
\centering
\includegraphics[width=\textwidth,keepaspectratio]{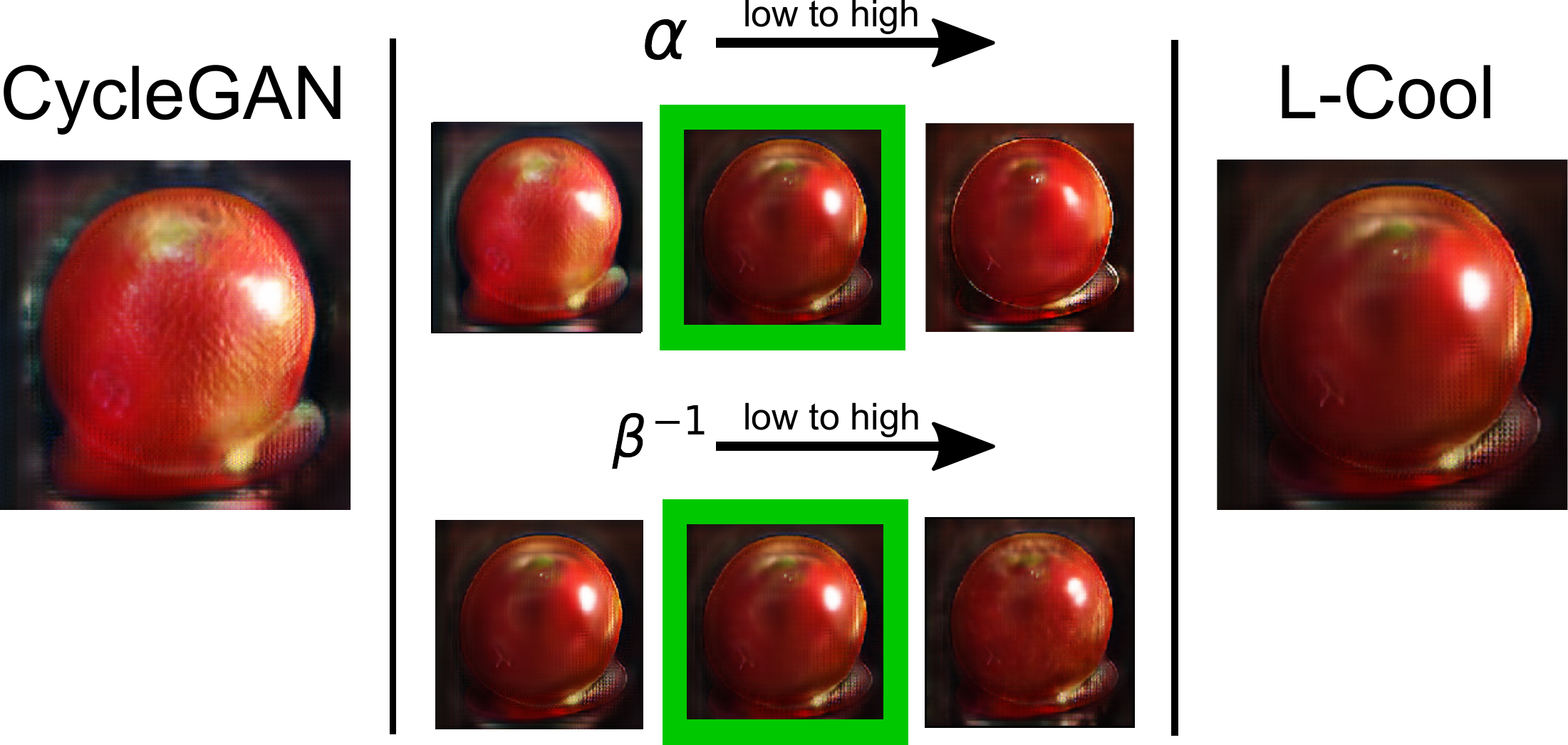}
\caption{
Translated images by \method{} with different 
hyperparameter settings.
We found that the setting $\beta^{-1} = 0.001$, $\alpha = 0.005$, and $N = 100$
(marked with a \emph{green} bounding box) 
best removes artifacts and increases the target domain attributes.
}
\label{fig:temp_o2a_1}
\end{figure*}

Table~\ref{table:sat2maps} shows the average pixel-wise accuracy, where 
we observed a similar tendency to the likeness evaluation in Section~\ref{sec:LikenessEvaluation}: for smaller proportions of fringe samples, the translation performance of the plain CycleGAN is worse, and the performance gain by \method{} is larger.
Figure~\ref{fig:sat2maps} shows an exemplar case
where \method{} improves translation performance.

\begin{figure}[t]
\centering

        \includegraphics[width=\columnwidth]{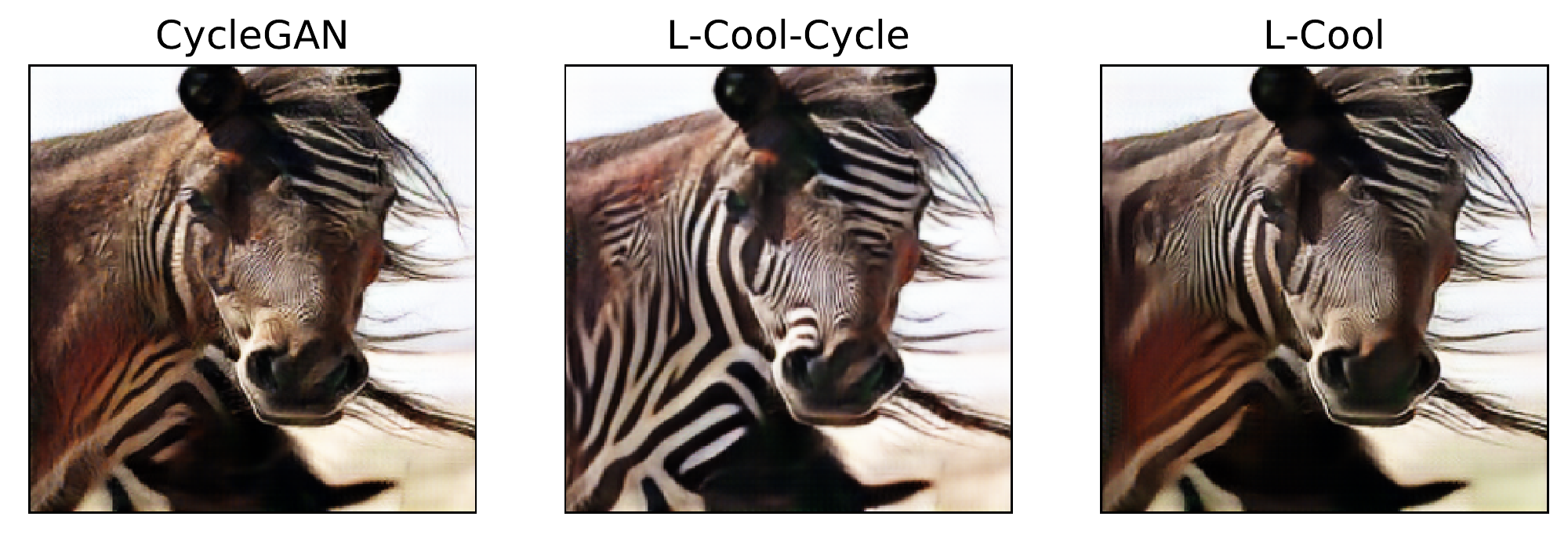}
    ~ \\
     \includegraphics[width=\columnwidth]{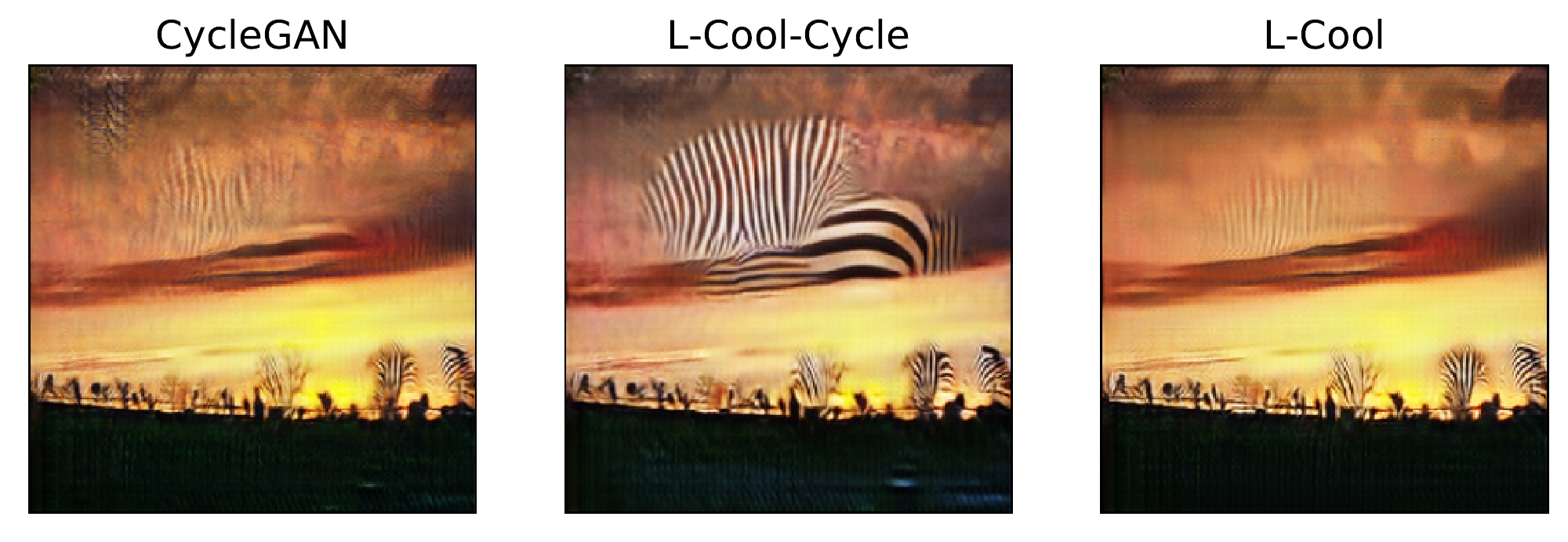}
    \caption{
Translated images by CycleGAN (left column), \method{}-Cycle (middle column), and \method{} (right column).  
\method{}-Cycle tends to enhance target domain attributes more than \method{} (top row), but also tends to exacerbate artifacts (bottom row). 
}
\label{fig:dae}
\end{figure}

\subsection{Hyperparameter Setting}
\label{sec:temp_analysis_image}

\method{} has several hyperparameters.
For DAE traning, 
we set the training noise to $\sigma = 0.3$ for all tasks,
which approximately follows the recommendation ($10\%$ of the mean pixel values) in \cite{Nguyen}.
We visually inspected the performance dependence on the remaining hyperparameters, i.e., 
temperature $\beta^{-1}$, 
step size $\alpha$, 
and the number of steps $N$.
Roughly speaking, the product of $\alpha$ and $N$ determines how far the resulting image can reach from the original point,
and similar results are obtained if $\alpha \cdot N$ has similar values, as long as the step size $\alpha$ is sufficiently small.

Figure~\ref{fig:temp_o2a_1} shows exemplarily translated images in the orange2apple task,
where the dependence on the temperature $\beta^{-1}$ and 
the step size $\alpha$ is shown for the number of steps fixed to $N = 100$.
We observed that, as the step size $\alpha$ increases, the translated image gets more attributes---increased red color on the apple---of the target domain, and artifacts are reduced.
However, if $\alpha$ is too large, the image gets blurred. 
We also observed that too high temperature $\beta^{-1}$ gives noisy result.
The visually best result 
was obtained when 
$\beta^{-1} = 0.001$, $\alpha = 0.005$ and $N = 100$ (marked with a green box and plotted on the right most in Figure~\ref{fig:temp_o2a_1}). 
Similar tendency was observed in other test samples and other tasks.
For quantitative evaluations in Section~\ref{sec:Image.Quantitative.Evaluation},
we optimized the hyperparameters on the validation set.
The reported results
were obtained with the hyperparameters searched over $\beta^{-1} = 0.0001, 0.001, 0.005, 0.01$, 
$\alpha = 0.001, 0.005, 0.01$, 
and
$N = 20, 40, 60, 80, 100$.




\begin{figure}[t]
\centering
\includegraphics[width=0.5\textwidth]{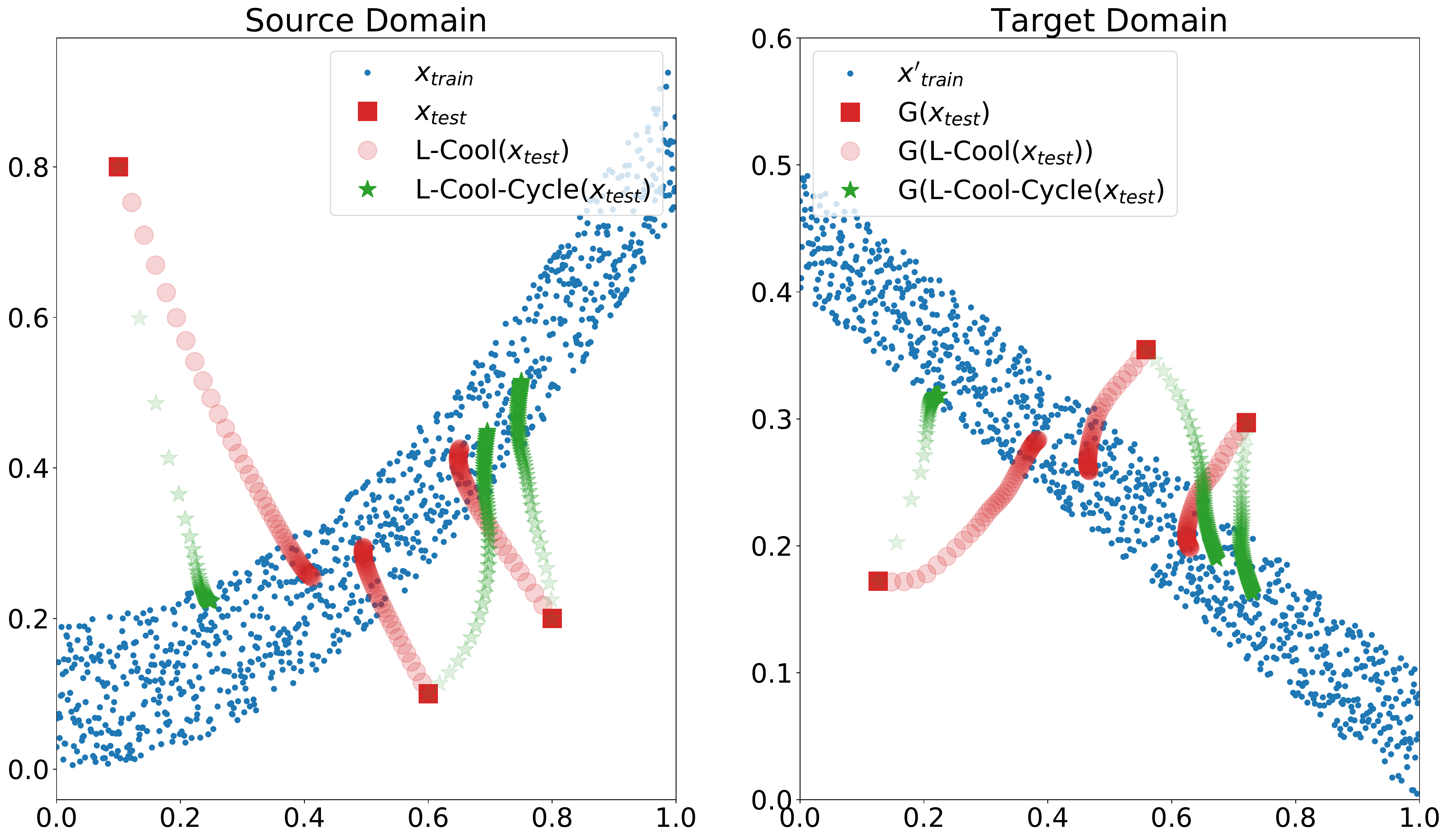}
\caption{
The same
toy data demonstration as Figure~\ref{fig:toy_data},
comparing
\method{} (red) and  \method{}-Cycle (green). 
In contrast with \method{},
 \method{}-Cycle does not move samples directly towards the high density region in the source domain, implying that the {cycle} gradient estimator is not a very good substitution for DAE gradient estimator.
}
\label{fig:toy_data_cycle}
\end{figure}


\begin{figure*}
\centering
\includegraphics[width=\textwidth]{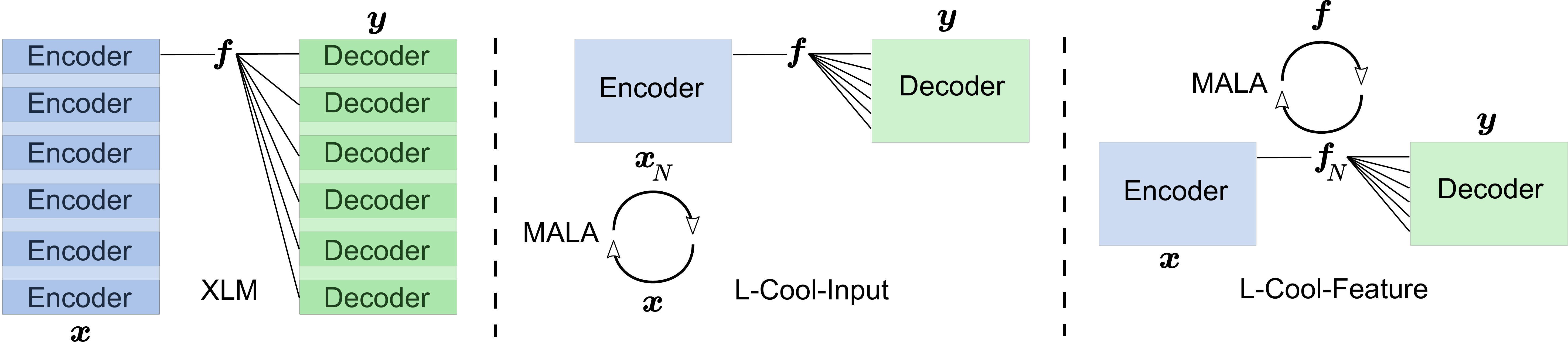}
\caption{
Schematics of XLM (left), \method{}-Input (middle), and \method{}-Feature (right).
\method{}-Input performs MALA in the input space, while 
\method{}-Feature performs MALA in the feature (code) space between the encoder and the decoder.
}
\label{fig:xlm_architecture}
\end{figure*}



\subsection{Investigation on the \method{}-Cycle}
\label{sec:grad_estimation}

\method{} requires a trained DAE for gradient estimation.  However, a variant, introduced in Section~\ref{sec:lcool_cycle} as an option called \method{}-Cycle,
eliminates the necessity of DAE training,
and estimate the gradient by using
the autoencoding structure of CycleGAN.
This option empirically showed good performance in image generation \cite{Nguyen}, as well as in our preliminary experiments 
in image translation
\cite{SriAAAI20}.

However, further investigation revealed a drawback of this variant: although \method{}-Cycle tends to enhance attributes of the target domain images, it also tends to exacerbate artifacts.
Figure~\ref{fig:dae} shows this tendency: \method{}-Cycle increases the contrast of stripes on the zebra body in the horse2zebra task (top row),
while it aggravates the stripe artifacts on the sky (bottom row).
In the latter case, we see that \method{} (with DAE) rather suppresses the artifacts.

Suboptimality of \method{}-Cycle
can already be seen in the toy data experiment.  Figure~\ref{fig:toy_data_cycle}
shows the same demonstration as in Figure~\ref{fig:toy_data},
and compares trails by \method{} and 
\method{}-Cycle.  
We see that 
\method{} (red) drives the off-manifold samples directly 
towards the data manifold, 
while
\method{}-Cycle (green) does not always do so.
This implies that the cycle estimator Eq.\eqref{eq:CycleGANEstimator} is not a very good gradient estimator.

In summary, although \method{}-Cycle is an option when training DAE is hard or time-consuming, 
it should be used in care---resulting samples should be checked by human.

\section{Language Translation Experiments}
\label{sec:UNMT_description}

In this section, 
we demonstrate the performance of 
our proposed \method{} in 
language translation tasks
with 
Cross-lingual Language Model (XLM) \cite{lample2017unsupervised, lample2019cross}---%
a state-of-the-art method
for unsupervised language translation---%
as the base method.


\subsection{Translation Tasks and Model Architectures}
\label{sec:TranslationTasks}

We conducted experiments
on four language translation tasks,
EN-FR, FR-EN, EN-DE, and DE-EN, based on NewsCrawl dataset\footnote{http://www.statmt.org/wmt14/}
under the default setting defined in the GitHub repository page:%
\footnote{https://github.com/facebookresearch/XLM
}
for each pair of languages, we used the first $5$M sentences for training, $3000$ sentences for validation, and $3000$ sentences for test.%

The main idea of XLM
is to share sub-word vocabulary 
between the source and the target languages created through 
the Byte Pair Encoding (BPE). 
Masked Language Modeling (MLM) is performed 
as pretraining, similarly to BERT \cite{devlin2018bert}. 
$15\%$ of the BPE from the text stream 
is masked  $80\%$ of the time, 
by a random token $10\%$ of the time
and they are kept unchanged $10\%$ of the time.
The encoder 
is pretrained with the MLM objective, 
whose weights are then used as 
initialization for both the encoder and the decoder.
This pretraining strategy 
was shown to give the best results \cite{lample2019cross}. 

The transformer consists of 
$6$ encoders and $6$ decoders.
The architectures of encoders and decoders
are similar, and each consists of 
a multi-head attention layer followed by 
layer normalization \cite{ba2016layer}, $2$ fully connected layers with GELU activations  \cite{hendrycks2016bridging} and another layer normalization. 
While the first fully connected layer projects the input with a 
dimensionality of $1024$
to a latent dimension of $4096$, the second fully connected layer
projects it back to $1024$. 
Each encoder and decoder layer 
also consists of 
a residual connection. 
For XLM implementation, 
we use
the code publicly available at
the GitHub page.
We train the model by using the ADAM optimizer along with linear warm-up and linear learning rates. 
We warm start with  
the model weights
obtained after the MLM stage,
and further train the weights on the training sentences.




We tested two variants of \method{} (see Figure~\ref{fig:xlm_architecture}).
\begin{description}
    \item [\method{}-Input:] 
    MALA is performed in the input word embedding space
    (the position embeddings are unaffected). 
    \item [\method{}-Feature:] 
MALA is performed in the intermediate feature (code) space. 
\end{description}
DAE with the same architecture as the encoder of the transformer
was trained in the corresponding space
on the training sentences 
of NewsCrawl.
Hyperparameters were tuned on the validation sentences (see Section~\ref{sec:temp_analysis_xlm}).

\begin{table}[t!]
\caption{BLEU scores in language translation tasks. 
}
\label{table:UNMT}
\resizebox{\columnwidth}{!}{

\begin{tabular}{c c c c c}
    \toprule
     & \textbf{EN-FR} & \textbf{FR-EN} & \textbf{EN-DE}& \textbf{DE-EN}\\
    \midrule
    XLM (Baseline) &33.46 & 31.62 & 25.51 & 31.11\\
    \method{}-Input & 31.59 & 31.90 & 25.66 & 30.93 \\
    \method{}-Feature & \textbf{33.91} & \textbf{31.93} & \textbf{25.73} & \textbf{31.17} \\
    \bottomrule    
    \end{tabular}
    }
\end{table}

\begin{figure*}[t]
\centering
 
    \begin{subfigure}[t]{\columnwidth}
        \centering
        \includegraphics[width=\columnwidth, 
        keepaspectratio]{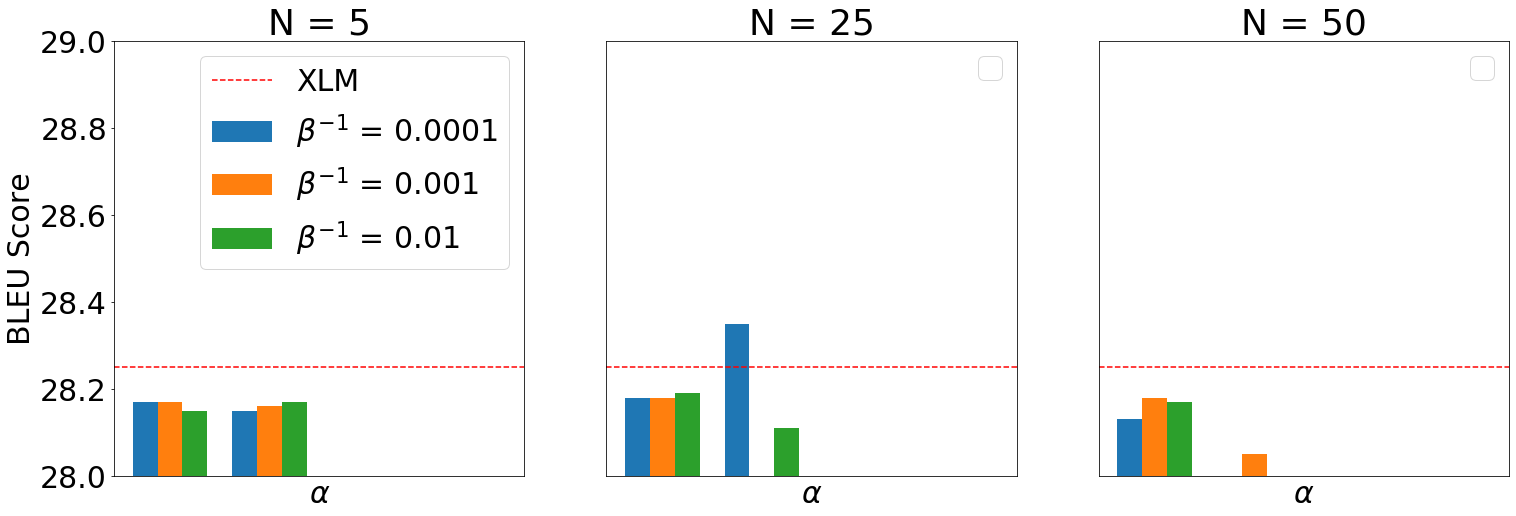}
        \caption{\method{}-Input}
        \label{fig:temp_xlm_inp}
        
    \end{subfigure}
    ~ 
    \begin{subfigure}[t]{\columnwidth}
        \centering
        \includegraphics[width=\columnwidth, keepaspectratio]{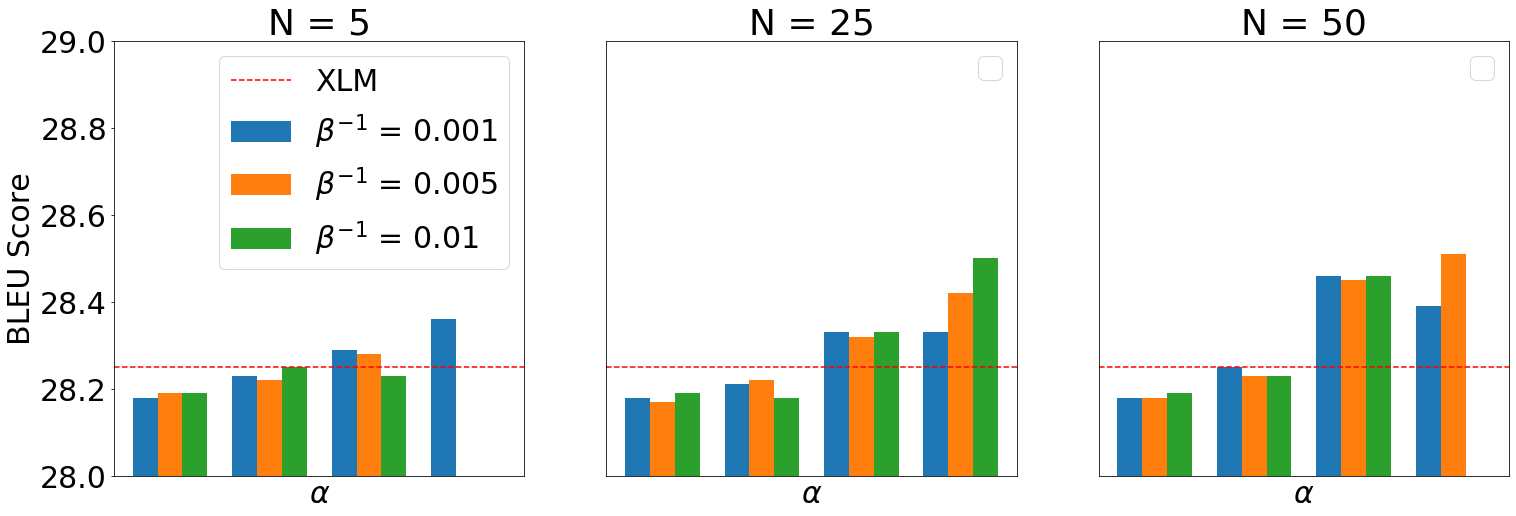}
        \caption{\method{}-Feature}
        \label{fig:temp_xlm_feat}
    \end{subfigure}
    ~
\caption{
Language translation performance (BLEU score) dependence on hyperparameters in the EN-FR task with \method{}-Input (left) and \method{}-Feature (right).
The dashed line in each graph indicates the baseline performance by plain XLM.
}
\label{fig:temp_xlm}
\end{figure*}

\subsection{Quantitative Evaluation}
\label{sec:Language.Quantitative.Evaluation}

Table~\ref{table:UNMT} shows the BLEU scores \cite{papineni2002bleu}
by plain XLM, \method{}-Input, and \method{}-Feature,
where we see consistent performance gain over all tasks by \method{}-Feature.
\method{}-Input does not improves 
the performance,
and even degrades in some tasks.
We conjecture that this is because of the discrete nature of the input space---the input is the word embedding that depends only on discrete occurrences of words, and therefore, a single step of MALA to any direction can bring the sample to a point where the base transformer is less trained than the original point.  

\subsection{Hyperparameter Setting}
\label{sec:temp_analysis_xlm}

Similarly to Section~\ref{sec:temp_analysis_image},
we set the DAE training noise to $\sigma^2 = 0.1$ for \method{}-Input and $\sigma^2 = 1.0$ for \method{}-Feature,
which approximately follow the recommendation in \cite{Nguyen}.
The remaining hyperparameters, i.e.,
temperature $\beta^{-1}$, 
step size $\alpha$, 
and the number of steps $N$,
were tuned 
by maximizing the BLEU score on the validation sentences.
The search ranges were $\beta^{-1} = 0.0001, 0.0005, 0.001, 0.005, 0.01$, 
$\alpha = 0.001, 0.005, 0.01$, $0.05$, $0.1$ and 
$N = 5, 25, 50$, respectively.

Figure~\ref{fig:temp_xlm}
shows performance dependence on the hyperparmaeters
for \method{}-Input (left) and 
\method{}-Feature (right) in the EN-FR translation task, 
where the best performance was obtained when
$\beta^{-1} = 10^{-4}, \alpha = 10^{-5}, N = 25 $ for \method{}-Input,
and when
$\beta^{-1} = 10^{-3}, \alpha=10^{-2}, N = 25 $ for \method{}-Feature.

\section{Computation Time}

\method{} requires additional computation cost both in training and test. Training the DAE 
can typically be done much faster than training the base DNN for the domain translation. 
In our experiment for the horse2zebra image translation task, training the DAE took $\sim 12800$ seconds or $3.55$ hours, while training the CycleGAN typically takes $\sim 42320$ seconds or $11.75$ hours (we did not train it because we used a pretrained network provided by the authors of CycleGAN). 
Note that 
this additional training is not necessary for \method{}-Cycle, which substitutes the cycle structure of the base DNN for gradient estimation.
In the test time,
\method{} requires $10$ to $100$ times more computation time, depending on the number of MALA steps.  This is because DAE should have a similar structure and complexity to the base DNN.
In our image translation experiment, \method{}
and CycleGAN took $\sim 5.3$ seconds and $\sim 0.5$ seconds per test image, respectively, while
in the language translation experiment,
\method{}
and XLM took 
$\sim 0.047$ seconds and $\sim 0.013$ per test sentence, respectively.


\section{Conclusion}

Developing unsupervised, as well as  self-supervised, learning methods, is one of the recent hot topics in the machine learning community for computer vision \cite{chen2020simple, he2020momentum, grill2020bootstrap, patrick2020multi, chen2020improved}
and natural language processing  \cite{devlin2018bert, xie2019unsupervised, yang2019xlnet, radford2019language, shoeybi2019megatron}.
It is challenging but highly attractive since eliminating the necessity of labeled data may enable us to keep improving learning machines from data stream automatically without any human intervention.
The successes of deep learning in the unsupervised domain translation (DT) was a milestone in this exciting research area.

Our work contributes to this area with a simple idea.  Namely, Langevin Cooling (\method{}) performs Metropolis Adjusted Langevin Algorithm (MALA) to test samples in the source domain, and drives them towards high density manifold, where the base deep neural network is well-trained.
Our qualitative and quantitative evaluations
showed improvements by \method{} in image and language translation tasks,
supporting our hypothesis that a proportion of test samples are failed to be translated because they lie at the fringe of data distribution, and therefore can be improved by \method{}.


\method{} is generic and can be used to improve any DT method.
Future work is therefore to apply \method{} to other base DT methods and other DT tasks.
We will also try to improve the gradient estimator for \method{} by using other types of generative models such as 
normalizing flows \cite{nfreview}.
Explanation methods, such as layer-wise relevance propagation (e.g. \cite{bach2015pixel,montavon2018methods,samek2020toward}), might help identify the reasons for successes and failures \cite{lapuschkin2019unmasking} of DT,
suggesting possible ways to improve the performance.


\section{Acknowledgements}
The authors acknowledge financial support  by  the  German  Ministry  for  Education  and  Research (BMBF) for the Berlin Center for Machine Learning  (01IS18037A), Berlin Big Data Center (01IS18025A) and  under  the  Grants  01IS14013A-E,  01GQ1115  and  01GQ0850;  Deutsche  Forschungsgemeinschaft  (DFG)  under  Grant  Math+,  EXC  2046/1,Project ID 390685689 and by the Technology Promotion(IITP) grant funded by the Korea government (No. 2017-0-00451).   Correspondence to WS, SN and KRM.

\bibliographystyle{IEEEtran} 
\bibliography{bibliography}

\begin{thebibliography}{10}
\providecommand{\url}[1]{#1}
\csname url@samestyle\endcsname
\providecommand{\newblock}{\relax}
\providecommand{\bibinfo}[2]{#2}
\providecommand{\BIBentrySTDinterwordspacing}{\spaceskip=0pt\relax}
\providecommand{\BIBentryALTinterwordstretchfactor}{4}
\providecommand{\BIBentryALTinterwordspacing}{\spaceskip=\fontdimen2\font plus
\BIBentryALTinterwordstretchfactor\fontdimen3\font minus
  \fontdimen4\font\relax}
\providecommand{\BIBforeignlanguage}[2]{{%
\expandafter\ifx\csname l@#1\endcsname\relax
\typeout{** WARNING: IEEEtran.bst: No hyphenation pattern has been}%
\typeout{** loaded for the language `#1'. Using the pattern for}%
\typeout{** the default language instead.}%
\else
\language=\csname l@#1\endcsname
\fi
#2}}
\providecommand{\BIBdecl}{\relax}
\BIBdecl

\bibitem{biswas2019prestack}
R.~Biswas, M.~K. Sen, V.~Das, and T.~Mukerji, ``Prestack and poststack
  inversion using a physics-guided convolutional neural network,''
  \emph{Interpretation}, vol.~7, no.~3, pp. SE161--SE174, 2019.

\bibitem{gilmer2017neural}
J.~Gilmer, S.~S. Schoenholz, P.~F. Riley, O.~Vinyals, and G.~E. Dahl, ``Neural
  message passing for quantum chemistry,'' in \emph{International Conference on
  Machine Learning}, 2017, pp. 1263--1272.

\bibitem{schutt2017schnet}
K.~Sch{\"u}tt, P.-J. Kindermans, H.~E.~S. Felix, S.~Chmiela, A.~Tkatchenko, and
  K.-R. M{\"u}ller, ``Schnet: A continuous-filter convolutional neural network
  for modeling quantum interactions,'' in \emph{Advances in neural information
  processing systems}, 2017, pp. 991--1001.

\bibitem{schutt2019unifying}
K.~Sch{\"u}tt, M.~Gastegger, A.~Tkatchenko, K.-R. M{\"u}ller, and R.~J. Maurer,
  ``Unifying machine learning and quantum chemistry with a deep neural network
  for molecular wavefunctions,'' \emph{Nature communications}, vol.~10, no.~1,
  pp. 1--10, 2019.

\bibitem{zhang2018deep}
G.~Zhang, Z.~Wang, and Y.~Chen, ``Deep learning for seismic lithology
  prediction,'' \emph{Geophysical Journal International}, vol. 215, no.~2, pp.
  1368--1387, 2018.

\bibitem{schmidt2019recent}
J.~Schmidt, M.~R. Marques, S.~Botti, and M.~A. Marques, ``Recent advances and
  applications of machine learning in solid-state materials science,''
  \emph{npj Computational Materials}, vol.~5, no.~1, pp. 1--36, 2019.

\bibitem{arridge2019solving}
S.~Arridge, P.~Maass, O.~{\"O}ktem, and C.-B. Sch{\"o}nlieb, ``Solving inverse
  problems using data-driven models,'' \emph{Acta Numerica}, vol.~28, pp.
  1--174, 2019.

\bibitem{bubba2019learning}
T.~A. Bubba, G.~Kutyniok, M.~Lassas, M.~M{\"a}rz, W.~Samek, S.~Siltanen, and
  V.~Srinivasan, ``Learning the invisible: A hybrid deep learning-shearlet
  framework for limited angle computed tomography,'' \emph{Inverse Problems},
  vol.~35, no.~6, p. 064002, 2019.

\bibitem{codevilla2018end}
F.~Codevilla, M.~Miiller, A.~L{\'o}pez, V.~Koltun, and A.~Dosovitskiy,
  ``End-to-end driving via conditional imitation learning,'' in \emph{2018 IEEE
  International Conference on Robotics and Automation (ICRA)}.\hskip 1em plus
  0.5em minus 0.4em\relax IEEE, 2018, pp. 1--9.

\bibitem{dosovitskiy2017carla}
A.~Dosovitskiy, G.~Ros, F.~Codevilla, A.~Lopez, and V.~Koltun, ``Carla: An open
  urban driving simulator,'' in \emph{Conference on Robot Learning}, 2017, pp.
  1--16.

\bibitem{AmazonCars}
\BIBentryALTinterwordspacing
Unknown, ``Developers, start your engines,'' 2020. [Online]. Available:
  \url{https://aws.amazon.com/deepracer/}
\BIBentrySTDinterwordspacing

\bibitem{VoyageDeepDrive}
\BIBentryALTinterwordspacing
D.~Gray, ``Introducing voyage deepdrive,'' 2019. [Online]. Available:
  \url{https://news.voyage.auto/introducing-voyage-deepdrive-69b3cf0f0be6}
\BIBentrySTDinterwordspacing

\bibitem{GoogleFederatedLearning}
\BIBentryALTinterwordspacing
B.~McMahan and D.~Ramage, ``Federated learning: Collaborative machine learning
  without centralized training data,'' 2017. [Online]. Available:
  \url{https://ai.googleblog.com/2017/04/federated-learning-collaborative.html-g/}
\BIBentrySTDinterwordspacing

\bibitem{wu2016google}
Y.~Wu, M.~Schuster, Z.~Chen, Q.~V. Le, M.~Norouzi, W.~Macherey, M.~Krikun,
  Y.~Cao, Q.~Gao, K.~Macherey \emph{et~al.}, ``Google's neural machine
  translation system: Bridging the gap between human and machine translation,''
  \emph{arXiv preprint arXiv:1609.08144}, 2016.

\bibitem{MicrosoftGaming}
\BIBentryALTinterwordspacing
Unknown, ``Game intelligence,'' 2020. [Online]. Available:
  \url{https://www.microsoft.com/en-us/research/theme/game-intelligence/}
\BIBentrySTDinterwordspacing

\bibitem{johnson2016perceptual}
J.~Johnson, A.~Alahi, and L.~Fei-Fei, ``Perceptual losses for real-time style
  transfer and super-resolution,'' in \emph{European conference on computer
  vision}.\hskip 1em plus 0.5em minus 0.4em\relax Springer, 2016, pp. 694--711.

\bibitem{he2016dual}
D.~He, Y.~Xia, T.~Qin, L.~Wang, N.~Yu, T.-Y. Liu, and W.-Y. Ma, ``Dual learning
  for machine translation,'' in \emph{Advances in neural information processing
  systems}, 2016, pp. 820--828.

\bibitem{lample2019cross}
A.~Conneau and G.~Lample, ``Cross-lingual language model pretraining,'' in
  \emph{Advances in Neural Information Processing Systems}, 2019, pp.
  7059--7069.

\bibitem{edunov2018understanding}
S.~Edunov, M.~Ott, M.~Auli, and D.~Grangier, ``Understanding back-translation
  at scale,'' \emph{Proceedings of the 2018 Conference on Empirical Methods in
  Natural Language Processing}, pp. 489--500, 2018.

\bibitem{gatys2015neural}
L.~A. Gatys, A.~S. Ecker, and M.~Bethge, ``A neural algorithm of artistic
  style,'' \emph{arXiv preprint arXiv:1508.06576}, 2015.

\bibitem{dong2015image}
C.~Dong, C.~C. Loy, K.~He, and X.~Tang, ``Image super-resolution using deep
  convolutional networks,'' \emph{IEEE transactions on pattern analysis and
  machine intelligence}, vol.~38, no.~2, pp. 295--307, 2015.

\bibitem{isola2017image}
P.~Isola, J.-Y. Zhu, T.~Zhou, and A.~A. Efros, ``Image-to-image translation
  with conditional adversarial networks,'' in \emph{Proceedings of the IEEE
  conference on computer vision and pattern recognition}, 2017, pp. 1125--1134.

\bibitem{ulyanov2018deep}
D.~Ulyanov, A.~Vedaldi, and V.~Lempitsky, ``Deep image prior,'' in
  \emph{Proceedings of the IEEE Conference on Computer Vision and Pattern
  Recognition}, 2018, pp. 9446--9454.

\bibitem{reed2016generative}
S.~Reed, Z.~Akata, X.~Yan, L.~Logeswaran, B.~Schiele, and H.~Lee, ``Generative
  adversarial text to image synthesis,'' \emph{arXiv preprint
  arXiv:1605.05396}, 2016.

\bibitem{zhang2017stackgan}
H.~Zhang, T.~Xu, H.~Li, S.~Zhang, X.~Wang, X.~Huang, and D.~N. Metaxas,
  ``Stackgan: Text to photo-realistic image synthesis with stacked generative
  adversarial networks,'' in \emph{Proceedings of the IEEE international
  conference on computer vision}, 2017, pp. 5907--5915.

\bibitem{sandfort2019data}
V.~Sandfort, K.~Yan, P.~J. Pickhardt, and R.~M. Summers, ``Data augmentation
  using generative adversarial networks (cyclegan) to improve generalizability
  in ct segmentation tasks,'' \emph{Scientific reports}, vol.~9, no.~1, pp.
  1--9, 2019.

\bibitem{wu2018conditional}
E.~Wu, K.~Wu, D.~Cox, and W.~Lotter, ``Conditional infilling gans for data
  augmentation in mammogram classification,'' in \emph{Image Analysis for
  Moving Organ, Breast, and Thoracic Images}.\hskip 1em plus 0.5em minus
  0.4em\relax Springer, 2018, pp. 98--106.

\bibitem{frid2018synthetic}
M.~Frid-Adar, E.~Klang, M.~Amitai, J.~Goldberger, and H.~Greenspan, ``Synthetic
  data augmentation using gan for improved liver lesion classification,'' in
  \emph{2018 IEEE 15th international symposium on biomedical imaging (ISBI
  2018)}.\hskip 1em plus 0.5em minus 0.4em\relax IEEE, 2018, pp. 289--293.

\bibitem{bowles2018gan}
C.~Bowles, L.~Chen, R.~Guerrero, P.~Bentley, R.~Gunn, A.~Hammers, D.~A. Dickie,
  M.~V. Hern{\'a}ndez, J.~Wardlaw, and D.~Rueckert, ``Gan augmentation:
  Augmenting training data using generative adversarial networks,'' \emph{arXiv
  preprint arXiv:1810.10863}, 2018.

\bibitem{goodfellow2014generative}
I.~Goodfellow, J.~Pouget-Abadie, M.~Mirza, B.~Xu, D.~Warde-Farley, S.~Ozair,
  A.~Courville, and Y.~Bengio, ``Generative adversarial nets,'' in
  \emph{Advances in neural information processing systems}, 2014, pp.
  2672--2680.

\bibitem{zhu2017unpaired}
J.-Y. Zhu, T.~Park, P.~Isola, and A.~A. Efros, ``Unpaired image-to-image
  translation using cycle-consistent adversarial networks,'' in
  \emph{Proceedings of the IEEE international conference on computer vision},
  2017, pp. 2223--2232.

\bibitem{kim2017learning}
T.~Kim, M.~Cha, H.~Kim, J.~K. Lee, and J.~Kim, ``Learning to discover
  cross-domain relations with generative adversarial networks,'' in
  \emph{Proceedings of the 34th International Conference on Machine
  Learning-Volume 70}.\hskip 1em plus 0.5em minus 0.4em\relax JMLR. org, 2017,
  pp. 1857--1865.

\bibitem{yi2017dualgan}
Z.~Yi, H.~Zhang, P.~Tan, and M.~Gong, ``Dualgan: Unsupervised dual learning for
  image-to-image translation,'' in \emph{Proceedings of the IEEE international
  conference on computer vision}, 2017, pp. 2849--2857.

\bibitem{vaswani2017attention}
A.~Vaswani, N.~Shazeer, N.~Parmar, J.~Uszkoreit, L.~Jones, A.~N. Gomez,
  {\L}.~Kaiser, and I.~Polosukhin, ``Attention is all you need,'' in
  \emph{Advances in neural information processing systems}, 2017, pp.
  5998--6008.

\bibitem{cycleganfailurecases}
``Cyclegan,'' \url{https://github.com/junyanz/CycleGAN#failure-cases}.

\bibitem{mejjati2018unsupervised}
Y.~A. Mejjati, C.~Richardt, J.~Tompkin, D.~Cosker, and K.~I. Kim,
  ``Unsupervised attention-guided image-to-image translation,'' in
  \emph{Advances in Neural Information Processing Systems}, 2018, pp.
  3693--3703.

\bibitem{Alain14}
G.~Alain and Y.~Bengio, ``What regularized auto-encoders learn from the
  data-generating distribution,'' \emph{The Journal of Machine Learning
  Research}, vol.~15, no.~1, pp. 3563--3593, 2014.

\bibitem{SriAAAI20}
V.~Srinivasan, K.-R. M{\"u}ller, W.~Samek, and S.~Nakajima, ``Benign examples:
  Imperceptible changes can enhance image translation performance,'' in
  \emph{Proceedings of the Thirty-Fourth AAAI Conference on Artificial
  Intelligence}, 2020.

\bibitem{papineni2002bleu}
K.~Papineni, S.~Roukos, T.~Ward, and W.-J. Zhu, ``Bleu: a method for automatic
  evaluation of machine translation,'' in \emph{Proceedings of the 40th annual
  meeting on association for computational linguistics}.\hskip 1em plus 0.5em
  minus 0.4em\relax Association for Computational Linguistics, 2002, pp.
  311--318.

\bibitem{wang2018high}
T.-C. Wang, M.-Y. Liu, J.-Y. Zhu, A.~Tao, J.~Kautz, and B.~Catanzaro,
  ``High-resolution image synthesis and semantic manipulation with conditional
  gans,'' in \emph{Proceedings of the IEEE conference on computer vision and
  pattern recognition}, 2018, pp. 8798--8807.

\bibitem{liu2017unsupervised}
M.-Y. Liu, T.~Breuel, and J.~Kautz, ``Unsupervised image-to-image translation
  networks,'' in \emph{Advances in neural information processing systems},
  2017, pp. 700--708.

\bibitem{huang2018multimodal}
X.~Huang, M.-Y. Liu, S.~Belongie, and J.~Kautz, ``Multimodal unsupervised
  image-to-image translation,'' in \emph{Proceedings of the European Conference
  on Computer Vision (ECCV)}, 2018, pp. 172--189.

\bibitem{lee2018diverse}
H.-Y. Lee, H.-Y. Tseng, J.-B. Huang, M.~Singh, and M.-H. Yang, ``Diverse
  image-to-image translation via disentangled representations,'' in
  \emph{Proceedings of the European Conference on Computer Vision (ECCV)},
  2018, pp. 35--51.

\bibitem{ugatit}
\BIBentryALTinterwordspacing
J.~Kim, M.~Kim, H.~Kang, and K.~Lee, ``{U-GAT-IT:} unsupervised generative
  attentional networks with adaptive layer-instance normalization for
  image-to-image translation,'' \emph{CoRR}, vol. abs/1907.10830, 2019.
  [Online]. Available: \url{http://arxiv.org/abs/1907.10830}
\BIBentrySTDinterwordspacing

\bibitem{bahdanau2014neural}
D.~Bahdanau, K.~Cho, and Y.~Bengio, ``Neural machine translation by jointly
  learning to align and translate,'' \emph{arXiv preprint arXiv:1409.0473},
  2014.

\bibitem{artetxe2017unsupervised}
M.~Artetxe, G.~Labaka, E.~Agirre, and K.~Cho, ``Unsupervised neural machine
  translation,'' \emph{arXiv preprint arXiv:1710.11041}, 2017.

\bibitem{radford2018improving}
A.~Radford, K.~Narasimhan, T.~Salimans, and I.~Sutskever, ``Improving language
  understanding by generative pre-training,'' 2018.

\bibitem{radford2019language}
A.~Radford, J.~Wu, R.~Child, D.~Luan, D.~Amodei, and I.~Sutskever, ``Language
  models are unsupervised multitask learners,'' \emph{OpenAI Blog}, vol.~1,
  no.~8, p.~9, 2019.

\bibitem{devlin2018bert}
J.~Devlin, M.-W. Chang, K.~Lee, and K.~Toutanova, ``Bert: Pre-training of deep
  bidirectional transformers for language understanding,'' in \emph{Proceedings
  of the 2019 Conference of the North American Chapter of the Association for
  Computational Linguistics: Human Language Technologies, Volume 1 (Long and
  Short Papers)}, 2019, pp. 4171--4186.

\bibitem{lewis2019bart}
``Bart: Denoising sequence-to-sequence pre-training for natural language
  generation, translation, and comprehension,'' \emph{arXiv preprint
  arXiv:1910.13461}, 2019.

\bibitem{lample2017unsupervised}
G.~Lample, A.~Conneau, L.~Denoyer, and M.~Ranzato, ``Unsupervised machine
  translation using monolingual corpora only,'' \emph{arXiv preprint
  arXiv:1711.00043}, 2017.

\bibitem{poncelas2018investigating}
A.~Poncelas, D.~Shterionov, A.~Way, G.~M. d.~B. Wenniger, and P.~Passban,
  ``Investigating backtranslation in neural machine translation,'' \emph{arXiv
  preprint arXiv:1804.06189}, 2018.

\bibitem{lample2018phrase}
G.~Lample, M.~Ott, A.~Conneau, L.~Denoyer, and M.~Ranzato, ``Phrase-based \&
  neural unsupervised machine translation,'' \emph{arXiv preprint
  arXiv:1804.07755}, 2018.

\bibitem{heek2019bayesian}
J.~Heek and N.~Kalchbrenner, ``Bayesian inference for large scale image
  classification,'' \emph{arXiv preprint arXiv:1908.03491}, 2019.

\bibitem{wenzel2020good}
F.~Wenzel, K.~Roth, B.~S. Veeling, J.~{\'S}wi{\k{a}}tkowski, L.~Tran, S.~Mandt,
  J.~Snoek, T.~Salimans, R.~Jenatton, and S.~Nowozin, ``How good is the bayes
  posterior in deep neural networks really?'' \emph{arXiv preprint
  arXiv:2002.02405}, 2020.

\bibitem{parmar2018image}
N.~Parmar, A.~Vaswani, J.~Uszkoreit, {\L}.~Kaiser, N.~Shazeer, A.~Ku, and
  D.~Tran, ``Image transformer,'' \emph{International Conference on Machine
  Learning}, pp. 4055--4064, 2018.

\bibitem{dahl2017pixel}
R.~Dahl, M.~Norouzi, and J.~Shlens, ``Pixel recursive super resolution,'' in
  \emph{Proceedings of the IEEE international conference on computer vision},
  2017, pp. 5439--5448.

\bibitem{kingma2018glow}
D.~P. Kingma and P.~Dhariwal, ``Glow: Generative flow with invertible 1x1
  convolutions,'' in \emph{Advances in neural information processing systems},
  2018, pp. 10\,215--10\,224.

\bibitem{Nguyen}
A.~Nguyen, J.~Clune, Y.~Bengio, A.~Dosovitskiy, and J.~Yosinski, ``Plug \& play
  generative networks: Conditional iterative generation of images in latent
  space,'' in \emph{Proceedings of the IEEE Conference on Computer Vision and
  Pattern Recognition}, 2017, pp. 4467--4477.

\bibitem{Vincent08}
P.~Vincent, H.~Larochelle, Y.~Bengio, and P.-A. Manzagol, ``Extracting and
  composing robust features with denoising autoencoders,'' in \emph{Proceedings
  of the 25th international conference on Machine learning}.\hskip 1em plus
  0.5em minus 0.4em\relax ACM, 2008, pp. 1096--1103.

\bibitem{Bengio13}
Y.~Bengio, L.~Yao, G.~Alain, and P.~Vincent, ``Generalized denoising
  auto-encoders as generative models,'' in \emph{Advances in Neural Information
  Processing Systems}, 2013, pp. 899--907.

\bibitem{Roberts96}
G.~O. Roberts and R.~L. Tweedie, ``Exponential convergence of {L}angevin
  distributions and their discrete approximations,'' \emph{Bernoulli}, vol.~2,
  pp. 341--363, 1996.

\bibitem{Roberts98}
G.~O. Roberts and J.~S. Rosenthal, ``Optimal scaling of discrete approximations
  to {L}angevin diffusions,'' \emph{Journal of the Royal Statistical Society,
  Series B}, vol.~60, pp. 255--268, 1998.

\bibitem{srinivasan2018counterstrike}
V.~Srinivasan, A.~Marban, K.-R. M{\"u}ller, W.~Samek, and S.~Nakajima,
  ``Robustifying models against adversarial attacks by langevin dynamics,''
  \emph{arXiv preprint arXiv:1805.12017}, 2018.

\bibitem{ioffe2015batch}
S.~Ioffe and C.~Szegedy, ``Batch normalization: Accelerating deep network
  training by reducing internal covariate shift,'' in \emph{International
  Conference on Machine Learning}, 2015, pp. 448--456.

\bibitem{jegou2017one}
S.~J{\'e}gou, M.~Drozdzal, D.~Vazquez, A.~Romero, and Y.~Bengio, ``The one
  hundred layers tiramisu: Fully convolutional densenets for semantic
  segmentation,'' in \emph{Proceedings of the IEEE conference on computer
  vision and pattern recognition workshops}, 2017, pp. 11--19.

\bibitem{pytorch}
A.~Paszke, S.~Gross, F.~Massa, A.~Lerer, J.~Bradbury, G.~Chanan, T.~Killeen,
  Z.~Lin, N.~Gimelshein, L.~Antiga, A.~Desmaison, A.~Kopf, E.~Yang, Z.~DeVito,
  M.~Raison, A.~Tejani, S.~Chilamkurthy, B.~Steiner, L.~Fang, J.~Bai, and
  S.~Chintala, ``Pytorch: An imperative style, high-performance deep learning
  library,'' in \emph{Advances in Neural Information Processing Systems 32}.

\bibitem{huang2017densely}
G.~Huang, Z.~Liu, L.~Van Der~Maaten, and K.~Q. Weinberger, ``Densely connected
  convolutional networks,'' in \emph{Proceedings of the IEEE conference on
  computer vision and pattern recognition}, 2017, pp. 4700--4708.

\bibitem{vgg}
K.~Simonyan and A.~Zisserman, ``Very deep convolutional networks for
  large-scale image recognition,'' \emph{arXiv preprint arXiv:1409.1556}, 2014.

\bibitem{szegedy2016rethinking}
C.~Szegedy, V.~Vanhoucke, S.~Ioffe, J.~Shlens, and Z.~Wojna, ``Rethinking the
  inception architecture for computer vision,'' in \emph{Proceedings of the
  IEEE conference on computer vision and pattern recognition}, 2016, pp.
  2818--2826.

\bibitem{xie2017aggregated}
S.~Xie, R.~Girshick, P.~Doll{\'a}r, Z.~Tu, and K.~He, ``Aggregated residual
  transformations for deep neural networks,'' in \emph{Proceedings of the IEEE
  conference on computer vision and pattern recognition}, 2017, pp. 1492--1500.

\bibitem{zagoruyko2016wide}
S.~Zagoruyko and N.~Komodakis, ``Wide residual networks,'' \emph{arXiv preprint
  arXiv:1605.07146}, 2016.

\bibitem{imagenet_cvpr09}
J.~Deng, W.~Dong, R.~Socher, L.-J. Li, K.~Li, and L.~Fei-Fei, ``{ImageNet: A
  Large-Scale Hierarchical Image Database},'' in \emph{CVPR09}, 2009.

\bibitem{ba2016layer}
J.~L. Ba, J.~R. Kiros, and G.~E. Hinton, ``Layer normalization,'' \emph{arXiv
  preprint arXiv:1607.06450}, 2016.

\bibitem{hendrycks2016bridging}
D.~Hendrycks and K.~Gimpel, ``Bridging nonlinearities and stochastic
  regularizers with gaussian error linear units,'' 2016.

\bibitem{chen2020simple}
T.~Chen, S.~Kornblith, M.~Norouzi, and G.~Hinton, ``A simple framework for
  contrastive learning of visual representations,'' \emph{arXiv preprint
  arXiv:2002.05709}, 2020.

\bibitem{he2020momentum}
K.~He, H.~Fan, Y.~Wu, S.~Xie, and R.~Girshick, ``Momentum contrast for
  unsupervised visual representation learning,'' in \emph{Proceedings of the
  IEEE/CVF Conference on Computer Vision and Pattern Recognition}, 2020, pp.
  9729--9738.

\bibitem{grill2020bootstrap}
J.-B. Grill, F.~Strub, F.~Altch{\'e}, C.~Tallec, P.~H. Richemond,
  E.~Buchatskaya, C.~Doersch, B.~A. Pires, Z.~D. Guo, M.~G. Azar \emph{et~al.},
  ``Bootstrap your own latent: A new approach to self-supervised learning,''
  \emph{arXiv preprint arXiv:2006.07733}, 2020.

\bibitem{patrick2020multi}
M.~Patrick, Y.~M. Asano, R.~Fong, J.~F. Henriques, G.~Zweig, and A.~Vedaldi,
  ``Multi-modal self-supervision from generalized data transformations,''
  \emph{arXiv preprint arXiv:2003.04298}, 2020.

\bibitem{chen2020improved}
X.~Chen, H.~Fan, R.~Girshick, and K.~He, ``Improved baselines with momentum
  contrastive learning,'' \emph{arXiv preprint arXiv:2003.04297}, 2020.

\bibitem{xie2019unsupervised}
Q.~Xie, Z.~Dai, E.~Hovy, M.-T. Luong, and Q.~V. Le, ``Unsupervised data
  augmentation for consistency training,'' \emph{arXiv preprint
  arXiv:1904.12848}, 2019.

\bibitem{yang2019xlnet}
Z.~Yang, Z.~Dai, Y.~Yang, J.~Carbonell, R.~R. Salakhutdinov, and Q.~V. Le,
  ``Xlnet: Generalized autoregressive pretraining for language understanding,''
  in \emph{Advances in neural information processing systems}, 2019, pp.
  5753--5763.

\bibitem{shoeybi2019megatron}
M.~Shoeybi, M.~Patwary, R.~Puri, P.~LeGresley, J.~Casper, and B.~Catanzaro,
  ``Megatron-lm: Training multi-billion parameter language models using gpu
  model parallelism,'' \emph{arXiv preprint arXiv:1909.08053}, 2019.

\bibitem{nfreview}
G.~Papamakarios, E.~Nalisnick, D.~J. Rezende, S.~Mohamed, and
  B.~Lakshminarayanan, ``Normalizing flows for probabilistic modeling and
  inference,'' \emph{arXiv preprint arXiv:1912.02762}, 2019.

\bibitem{bach2015pixel}
S.~Bach, A.~Binder, G.~Montavon, F.~Klauschen, K.-R. M{\"u}ller, and W.~Samek,
  ``On pixel-wise explanations for non-linear classifier decisions by
  layer-wise relevance propagation,'' \emph{PloS one}, vol.~10, no.~7, p.
  e0130140, 2015.

\bibitem{montavon2018methods}
G.~Montavon, W.~Samek, and K.-R. M{\"u}ller, ``Methods for interpreting and
  understanding deep neural networks,'' \emph{Digital Signal Processing},
  vol.~73, pp. 1--15, 2018.

\bibitem{samek2020toward}
W.~Samek, G.~Montavon, S.~Lapuschkin, C.~J. Anders, and K.-R. M{\"u}ller,
  ``Toward interpretable machine learning: Transparent deep neural networks and
  beyond,'' \emph{arXiv preprint arXiv:2003.07631}, 2020.

\bibitem{lapuschkin2019unmasking}
S.~Lapuschkin, S.~W{\"a}ldchen, A.~Binder, G.~Montavon, W.~Samek, and K.-R.
  M{\"u}ller, ``Unmasking clever hans predictors and assessing what machines
  really learn,'' \emph{Nature communications}, vol.~10, no.~1, p. 1096, 2019.

\end{thebibliography}

\end{document}